\documentclass{article}

\usepackage{microtype}
\usepackage{graphicx}
\usepackage{subfigure}
\usepackage{booktabs}

\usepackage{hyperref}

\usepackage{booktabs}

\usepackage[accepted]{icml2023}

\usepackage{amsmath}
\usepackage{amssymb}
\usepackage{mathtools}
\usepackage{amsthm}
\usepackage[capitalize,noabbrev]{cleveref}

\theoremstyle{plain}

\theoremstyle{definition}

\theoremstyle{remark}

\usepackage[textsize=tiny]{todonotes}
\usepackage{multirow}
\usepackage[normalem]{ulem}
\usepackage{makecell}
\usepackage{enumitem}

\usepackage{colortbl}
\useunder{\uline}{\ul}{}

\usepackage[capitalize,noabbrev]{cleveref}

\icmltitlerunning{Dink-Net: Neural Clustering on Large Graphs}

\begin{document}

\twocolumn[
\icmltitle{Dink-Net: Neural Clustering on Large Graphs}

\icmlsetsymbol{Corresponding Author}{$\dag$}

\begin{icmlauthorlist}
\icmlauthor{Yue Liu}{NUDT,Westlake}
\icmlauthor{Ke Liang}{NUDT}
\icmlauthor{Jun Xia}{Westlake}
\icmlauthor{Sihang Zhou}{NUDT}
\icmlauthor{Xihong Yang}{NUDT}
\icmlauthor{Xinwang Liu}{NUDT,Corresponding Author}
\icmlauthor{Stan Z. Li}{Westlake,Corresponding Author}


\end{icmlauthorlist}

\icmlaffiliation{NUDT}{National University of Defense Technology}
\icmlaffiliation{Westlake}{Westlake University}

\icmlcorrespondingauthor{Yue Liu}{yueliu19990731@163.com}
\icmlcorrespondingauthor{Xinwang Liu}{xinwangliu@nudt.edu.cn}
\icmlcorrespondingauthor{Stan Z. Li}{Stan.ZQ.Li@westlake.edu.cn}


\icmlkeywords{Machine Learning, ICML}

\vskip 0.3in
]

\printAffiliationsAndNotice{\icmlEqualContribution}


\renewcommand{\thefootnote}{\fnsymbol{footnote}}

\begin{abstract}


\renewcommand{\thefootnote}{\Roman{footnote}}Deep graph clustering, which aims to group the nodes of a graph into disjoint clusters with deep neural networks, has achieved promising progress in recent years. However, the existing methods fail to scale to the large graph with million nodes. To solve this problem, a scalable deep graph clustering method (\textit{Dink-Net}) is proposed with the idea of \underline{di}lation and shri\underline{nk}. Firstly, by discriminating nodes, whether being corrupted by augmentations, representations are learned in a self-supervised manner. Meanwhile, the cluster centers are initialized as learnable neural parameters. Subsequently, the clustering distribution is optimized by minimizing the proposed cluster dilation loss and cluster shrink loss in an adversarial manner. By these settings, we unify the two-step clustering, i.e., representation learning and clustering optimization, into an end-to-end framework, guiding the network to learn clustering-friendly features. Besides, \textit{Dink-Net} scales well to large graphs since the designed loss functions adopt the mini-batch data to optimize the clustering distribution even without performance drops. Both experimental results and theoretical analyses demonstrate the superiority of our method. Compared to the runner-up, \textit{Dink-Net} achieves $9.62\%$ NMI improvement on the ogbn-papers100M dataset with 111~million nodes and 1.6~billion edges. The source code is released: \textit{Dink-Net}\footnote[1]{https://github.com/yueliu1999/Dink-Net}. Besides, a collection (papers, codes, and datasets) of deep graph clustering is shared on GitHub\footnote[2]{https://github.com/yueliu1999/Awesome-Deep-Graph-Clustering}. 

\end{abstract}


\section{Introduction}
Attribute graph clustering, which aims to separate the nodes in an attribute graph into different groups, has become a fast-growing research direction in recent years. As a pure unsupervised mission, promising achievements are made by models based on deep neural networks, especially graph neural networks (GNN) \cite{GCN,GAE,GAT}. Multiple models are proposed \cite{MGAE,MVGRL,AGCN,DCRN,AGC-DRR,S3GC}, which generally first embed nodes into the hidden space and then perform clustering algorithms on them. Although proven effective, most existing models in such a manner suffer from scalability issues, i.e., poor scalability on large graphs. Such a problem is one of the most critical challenging tasks in deep graph clustering \cite{liuyue_survey}, and our work attempts to seek a breakthrough according to it.


Based on our observations, there are three main reasons for the poor scalability of existing models on large graphs with millions of nodes. Firstly, some methods \cite{AGE,MVGRL,DCRN,AGC-DRR} need to process the $N \times N$ dense graph diffusion matrix \cite{graph_diffusion}, where $N$ is the number of nodes. Secondly, unlike other tasks like node classification or link prediction, clustering requires the method to estimate the whole sample distribution at once. Therefore, when the node number grows to a considerable value, e.g., 100 million, it easily leads to out-of-memory failure or long-running time problems. Thirdly, current attempts for this problem, such as S$^3$GC \cite{S3GC}, usually separate the optimization of representation learning and clustering, leading to sub-optimal performance.


To solve this problem, we present a scalable deep graph clustering method termed dilation shrink network (\textit{Dink-Net}). The guidance idea is motivated by the dilation and shrink of many galaxies in the universe. Concretely, it mainly consists of the node discriminate module and the neural clustering module. At first, in the node discriminate module, node representations are learned by telling apart the original samples and augmented samples. After the pre-training, the cluster centers are initialized and assigned as optimizable neural parameters with gradients. Additionally, at fine-tuning stage, the neural clustering module optimizes the clustering distribution by minimizing the proposed dilation and shrink loss functions. Concretely, in an adversarial manner, the dilation loss attempts to expand clustering distribution by pushing away different clusters, while the shrink loss aims to compress the clustering distribution by pulling samples back to cluster centers.

With the above designs, \textit{Dink-Net} learns the clustering-friendly representations via unifying the representation learning and clustering optimization into an end-to-end framework. In addition, the proposed loss functions effectively optimize clustering distribution on mini-batch data. Therefore, it endows the scalability of our method without performance drops. Notably, \textit{Dink-Net} is scaled on a large graph like ogbn-papers100M with 111 million nodes and 1.6 billion edges. The main contributions are summarized as follows. 

\begin{itemize}

    \item A scalable deep graph clustering method named dilation shrink network (\textit{Dink-Net}) is proposed to expand the deep graph clustering to large-scale graphs.

    \item We are the first to optimize the clustering distribution via the designed dilation and shrink loss functions in an adversarial manner. The method only relies on the mini-batch data, thus, endowing promising scalability.
    
    \item We unify the representation learning and clustering optimization procedures into an end-to-end framework for better clustering-friendly features, leading to superior clustering performance.


    \item Both experimental results and theoretical analyses are provided to verify the capability of \textit{Dink-Net} from six aspects, \textit{i.e.,} superiority, effectiveness, scalability, efficiency, sensitivity, and convergence.

\end{itemize}



\section{Related Work}

\subsection{Deep Graph Clustering}

Graph Neural Networks \cite{GCN,GAT,GAE,liuyixing_survey} become popular in different graph scenarios \cite{liang2022reasoning, liang2022relational, meng2023sarf, liang2023abslearn, liu2023self, liang2023message}. Among these, attribute graph clustering is a fundamental yet challenging task to separate the nodes in the attribute graph into different clusters without human annotations.

The early methods \cite{GraphEncoder,DNGR} adopt auto-encoders to learn node embeddings and then perform $K$-Means on them. Subsequently, motivated by the success of graph neural networks (GNNs) \cite{GAE,GCN}, MGAE \cite{MGAE} is proposed to encode nodes with graph-auto-encoders and then group the nodes into clusters with the spectral clustering algorithm. \cite{ARGA_conf} propose ARGA by enforcing the latent representations to align a prior distribution. To design a clustering-directed method, they propose a unified framework termed DAEGC \cite{DAEGC} with the attention-based graph encoder and clustering alignment loss adopted in deep clustering methods \cite{DEC}. SDCN \cite{SDCN} verifies the effectiveness of integrating structural and attribute information. Then, to avoid the expensive costs of spectral clustering, \cite{mincutpool} formulate a continuous relaxation of the normalized minCUT problem and optimize the clustering objective with the GNNs. More recently, the contrastive learning \cite{Mo_AAAI_2022,mo2023multiplex,zheng2022toward,zheng2022unifying} become hot research hot in deep graph clustering domain \cite{CCGC,GCC-LDA,ICRN}. Concretely, AGE \cite{AGE} filters the high-frequency noises in node attributes and then trains the encoder by adaptively contrasting the positive and negative samples. MVGRL \cite{MVGRL} generates an augmented structural view and contrasts node embeddings from one view with graph embeddings of another view and vice versa. Although the effectiveness of the contrastive learning paradigm is verified, there are still many open technical problems. Specifically, \cite{GDCL} proposes GDCL to correct sampling bias in contrastive deep graph clustering. Moreover, \cite{DCRN,IDCRN} design the dual correlation reduction strategy in the DCRN model to alleviate the representation collapse problem. Besides, HSAN \cite{liuyue_HSAN} mines the hard sample pairs via the dynamic weighting strategy. And SCGC \cite{liuyue_SCGC} simplifies the graph augmentation with parameter-unshared Siamese encoders and embedding disturbance. TGC \cite{TGC_liumeng} present a general framework for deep node clustering on temporal graphs. For more details about deep graph clustering, refer to the survey paper \cite{liuyue_survey}. However, most previous methods fail to scale to large graphs with millions of nodes. In order to alleviate this problem, a scalable deep graph clustering method termed S$^3$GC \cite{S3GC} is proposed by contrastive learning along with GNNs. Although verifying the effectiveness, they separate the optimization of representation learning and clustering, leading to sub-optimal performance. This paper presents a new scalable method that unifies embedding and clustering into an end-to-end framework. Therefore, our method not only scales to the large graphs but also learns the clustering-friendly representations.

\subsection{Salable Graph Neural Network}
Graph Neural Networks (GNNs) \cite{GCN,GAT} become one of the most effective tools for learning over graph data. Many scalable GNNs have been proposed to scale to large graphs in recent years. For example, GraphSAGE \cite{GraphSage} develops a general inductive framework by sampling and aggregating features from the local neighborhood of the nodes. FastGCN \cite{Fastgcn} avoids the recursive neighborhood expansion by the layer-wise sampling to nodes in each layer independently. Additionally, SGC \cite{SGC} decouples the transformation and propagation in GCN \cite{GCN}. Besides, Graphsaint \cite{Graphsaint} and Cluster-GCN \cite{Cluster-gcn} are proposed to better maintain graph structure by sub-graph sampling. Moreover, \cite{Deepgcns,Deepergcn,1000layers,liu2020towards} aim to design a sequence of works to make GCNs deeper. And various normalization and regularization techniques like DropEdge \cite{Dropedge} and ParNorm \cite{Pairnorm} are proposed to avoid over-fitting and over-smoothing. Furthermore, \cite{bojchevski2019pagerank,PPPGo,Sign} attempt to propose more efficient propagation schemes. More recently, \cite{Scalegcn} have designed a new efficient graph convolution via channel-wise scale transformation. \cite{GGD} scale up the graph contrastive learning by simplifying DGI \cite{DGI} and designing discriminate tasks. Sketch-GNN \cite{Sketch-GNN} is proposed by training GNNs atop a few compact sketches of graph structure and node features. At the same time, \cite{NodeFormer,rampavsek2022recipe} propose the scalable graph transformer model. However, the scalable GNNs for clustering tasks are few. It is challenging since the clustering task needs the model to estimate the whole sample distribution at once. Therefore, this paper aims to extend deep graph clustering methods to large-scale graphs.

\section{Methodology}
The methodology of \textit{Dink-Net} is introduced in this section. We first define the problem and summary the basic notation. Then, the challenges of scaling deep graph clustering methods to large graphs are carefully illustrated. In addition, our solution to this problem is provided with the reasons.

\subsection{Basic Notation}
Given an attribute graph $\boldsymbol{G}$, $\boldsymbol{V}=\{v_1,v_2,...v_N\}$ denotes a set of vertices, and $\boldsymbol{E} \subseteq \left\{(x,y) \middle|  (x,y)\in \boldsymbol{V}^2\right\}$ denotes a set of edges between vertices, where each vertex attaches the corresponding $D$-dimension attributes. $\textbf{X} \in \mathbb{R}^{N \times D}$ and $\textbf{A} \in \mathbb{R}^{N \times N}$ are defined as the node attribute matrix and adjacency matrix, separately. Here, $N$ and $D$ denote the number of vertices and dimension number of the attributes, respectively. The basic notation table is presented in Table~1 of Appendix.

\subsection{Problem Definition}
For an attribute graph $\boldsymbol{G}$, the deep graph clustering algorithm aims to group the vertices into disjoint clusters. Specifically, the self-supervised neural network $\mathcal{F}$ embeds the nodes in $\boldsymbol{G}$ into the latent space as follows.
\begin{equation} 
\textbf{H} = \mathcal{F}(\boldsymbol{G}) = \mathcal{F}(\textbf{X},\textbf{A}),
\label{deep_graph_clustering_encoding}
\end{equation}
where $\textbf{H} \in \mathbb{R}^{N \times d}$ denotes the node embeddings and $d$ is the dimension number of latent features. Here, the self-supervised network $\mathcal{F}$ is trained with the pre-text tasks like reconstructive task, contrastive task, discriminative task, etc. In addition to encoding, the clustering method $\mathcal{C}$ is designed to group the nodes into different clusters as follows. 
\begin{equation} 
\hat{\textbf{y}} = \mathcal{C}(\textbf{H}, K),
\label{deep_graph_clustering_clustering}
\end{equation}
where $K$ is the number of clusters, which can be a hyper-parameter or a learnable parameter in the clustering method $\mathcal{C}$. The result $\hat{\textbf{y}} \in \mathbb{R}^{N}$ is the clustering assignment vector.

Models suitable for large-scale graphs are always the goal researchers pursue. Unlike node classification and link prediction tasks, performing node clustering on a large-scale graph is more challenging. To this end, we aim to propose a method, which can empower the deep graph clustering algorithms to perform well on large-scale graphs, e.g., the graph with $\sim$111~million nodes and $\sim$10~billion edges. The detailed reasons are analyzed in the following sub-section.

\subsection{Challenge Analyses}
\label{sec:challenge}
This section carefully analyzes the challenges of large-scale deep graph clustering. It begins with the differences between deep graph clustering and other tasks like node classification and link prediction. For the node classification task, instead of processing the whole graph data at once, algorithms can divide data into mini-batches \cite{mini-batch_training} and merely classify samples in each mini-batch. Similarly, models adopt the mini-batch technique for the link prediction task for the paired nodes and predict the probability of the links between paired nodes in mini-batches. The mini-batch technique works because the predictions of each node or link are relatively independent, and they will not influence each other at the inference stage. However, in the node clustering task, the methods need to group all nodes into disjoint clusters at once. In this process, the cluster assignment of each node will influence each other, and therefore the mini-batch technique easily fails. 

Due to the above concerns, most existing deep graph clustering methods fail to use the mini-batch technique and process the whole data at once in the clustering process. Concretely, one class of methods first embeds nodes into the latent space and then directly performs the traditional clustering algorithm \cite{K-Means,spectral_clustering} on the learned node representations. We first analyze the complexity of traditional clustering methods. For example, the time complexity and space complexity of $K$-Means algorithm \cite{K-Means} is $\mathcal{O}(tNKD)$ and $\mathcal{O}(NK+ND+KD)$. Here, $t$, $N$, $K$, and $D$ denote the iteration times, node number, cluster number, and attribute dimension number, respectively. In addition, the time complexity and space complexity of spectral clustering algorithm \cite{spectral_clustering} is $\mathcal{O}(N^3)$ and $\mathcal{O}(N^2)$.

Besides, the above methods separate the embedding and clustering optimization process, leading to sub-optimal performance. Differently, another class of methods \cite{DAEGC,SDCN,DCRN} unify the representation learning and clustering optimization into a joint framework by minimizing the KL divergence loss \cite{DEC,IDEC} as follows. 
\begin{equation} 
\min \mathcal{L}_{\textrm{KL}} = \min \sum_i \sum_j \textbf{P}_{ij}log\left(\frac{\textbf{P}_{ij}}{\textbf{Q}_{ij}}\right),
\label{pq_loss}
\end{equation}
where $\textbf{Q}_{ij}$ is the original clustering distribution and $\textbf{P}_{ij}$ is the sharpened clustering distribution. This loss function optimizes the clustering distribution with the whole data. Thus, the calculation and optimization process is complex and resource-consuming, leading to $\mathcal{O}(NKd)$ time complexity and $\mathcal{O}(NK+Nd+Kd)$ space complexity.



Therefore, when the number of nodes $N$ reaches a considerable value like $\sim$111~million on the ogbn-papers100M dataset, the previous two types of methods lead to unacceptable running time and out-of-memory problems. Besides, some methods \cite{MVGRL,GDCL,DCRN,AGC-DRR} need to process the $N \times N$ dense graph diffusion matrix \cite{graph_diffusion}, which also hinders the efficiency. To this end, we develop a scalable end-to-end deep graph clustering method with the guidance of divide-and-rule.


\begin{algorithm}[h]
\footnotesize
\caption{Dilation Shrink Network (\textit{Dink-Net})}
\label{algorithm:dink-net}
\textbf{Input}: Attribute graph $\boldsymbol{G}$; cluster number $K$; epoch number $T,T'$; learning rate $\beta$,$\beta'$; batch size $B$; trade-off parameter $\alpha$.\\
\textbf{Output}: Predicted cluster-ID $\hat{\textbf{y}}$.
\begin{algorithmic}[1]
\STATE Initialize model parameters $\Theta$ in encoder $\mathcal{F}$ and projection $\mathcal{P}$;
\STATE $\#$ Model pre-training stage
\FOR{$\text{epoch}=1,2,...,T$}
\STATE Obtain new graph $\boldsymbol{G}' = \{\textbf{X}', \textbf{A}'\}$ via data augmentations;
\STATE Node encoding: $\textbf{H} = \mathcal{F}(\boldsymbol{G})$, $\textbf{H}' = \mathcal{F}(\boldsymbol{G}')$;
\STATE Representation projection: $\textbf{Z} = \mathcal{P}(\textbf{H})$, $\textbf{H}' = \mathcal{P}(\textbf{H}')$;
\STATE Node summary: $\textbf{g} = \textbf{Z}.\text{sum}(1)$, $\textbf{g}' = \textbf{Z}'.\text{sum}(1)$;
\STATE Calculate discrimination loss $\mathcal{L}_{\text{discri.}}$ in Eq. \eqref{eq:discriminative_loss};
\STATE Adam optimizer with learning rate $\beta$ updates parameters $\Theta$ by minimizing $\mathcal{L}_{\text{discri.}}$;
\ENDFOR

\STATE $\#$ Model fine-tuning stage
\STATE Initialize the cluster center embeddings $\textbf{C}$ in the K-means++ manner based on the learned node embeddings \textbf{H};
\FOR{$\text{epoch}=1,2,...,T'$}
\STATE Generate batched graph data $\boldsymbol{B}$ with shuffle;
\FOR{$\textbf{G}_{\boldsymbol{B}}$ in $\boldsymbol{B}$}
\STATE Node encoding: $\textbf{H}=\mathcal{F}(\textbf{G}_{\boldsymbol{B}})$;
\STATE Calculate discrimination loss $\mathcal{L}_{\text{discri.}}$ in Eq. \eqref{eq:discriminative_loss};
\STATE Calculate dilation loss $\mathcal{L}_{\text{dilation}}$ in Eq. \eqref{eq:dilation_loss};
\STATE Calculate shrink loss $\mathcal{L}_{\text{shrink}}$ in Eq. \eqref{eq:shrink_loss};
\STATE Calculate the total loss $\mathcal{L}$ in Eq. \eqref{eq:total_loss};
\STATE Adam optimizer with learning rate $\beta'$ updates parameters $\Theta$ and cluster centers $\textbf{C}$ by minimizing $\mathcal{L}$;
\ENDFOR

\ENDFOR
\STATE $\#$ Model inference stage
\STATE Generate batched graph data $\boldsymbol{B}$ without shuffle;
\FOR{$\boldsymbol{G}_{\boldsymbol{B}}$ in $\boldsymbol{B}$}
\STATE Node encoding: $\textbf{H}=\mathcal{F}(\boldsymbol{G}_{\boldsymbol{B}})$;
\STATE Predict cluster-ID of batch data $\hat{\textbf{y}}_{\boldsymbol{B}}$ by Eq. \eqref{eq:assignment};
\ENDFOR
\STATE Concatenate batched cluster-ID and obtain $\hat{\textbf{y}}$;
\STATE \textbf{Return} $\hat{\textbf{y}}$
\end{algorithmic}
\end{algorithm}













\subsection{Proposed Solution}
Through careful analyses in the previous section, we conclude that it is challenging to expand existing deep graph clustering methods to large-scale graphs. To solve this problem, we propose a scalable end-to-end method named dilation shrink network (\textit{Dink-Net}) with the idea of dilation and shrink of galaxies in the universe. Intuitively, our method optimizes the clustering distribution with the mini-batch data in an adversarial manner, therefore scaling to the large graph. \textit{Dink-Net} mainly comprises the following node discriminate module and neural clustering module.

\textbf{Node Discriminate Module.} Given an attribute graph $\boldsymbol{G}$, we first apply the graph data augmentations $\tau$, like attribute disturbance and edge dropout, on the node attributions $\textbf{X} \in \mathbb{R}^{N \times D}$ and graph structure $\textbf{A} \in \mathbb{R}^{N \times N}$. As the result, a augmented view graph view $\boldsymbol{G}'$ is constructed with $\textbf{X}'$ and $\textbf{A}'$. Subsequently, the parameter-share graph neural network encoder $\mathcal{F}(\cdot)$ embeds the nodes of $\boldsymbol{G},\boldsymbol{G}'$ to the latent embeddings $\textbf{H}, \textbf{H}' \in \mathbb{R}^{N \times d}$. Then, a small parameter-shared neural network projection head $\mathcal{P}(\cdot)$ maps the nodes embeddings into a new latent space, where the self-supervised learning loss will be applied. It outputs $\textbf{Z},\textbf{Z}' \in \mathbb{R}^{N \times d}$. After that, our method pools the new node embeddings into the node summaries $\textbf{g},\textbf{g}' \in \mathbb{R}^{N \times 1}$ by the feature aggregation operation. To train the encoder $\mathcal{F}(\cdot)$ and projection head $\mathcal{P}(\cdot)$, a binary cross entropy loss function is minimized to tell apart the original node and the augmented node summaries below.
\begin{equation} 
\begin{aligned}
\min \mathcal{L}_{\text{discri.}} = & \min \left[ \frac{1}{N} \sum_{i=1}^{N} \left(1 \cdot log\frac{1}{\textbf{g}_i} + 0 \cdot log\frac{1}{1-\textbf{g}_i}\right) + \right. \\ & \left. \frac{1}{N} \sum_{i=1}^{N} \left(0 \cdot log\frac{1}{\textbf{g}'_i} + 1 \cdot log\frac{1}{1-\textbf{g}'_i}\right) \right]  = \\ & \min \frac{1}{N} \sum_{i=1}^N\left(log\frac{1}{\textbf{g}_i}+log\frac{1}{1-\textbf{g}_i'}\right).     
\end{aligned}
\label{eq:discriminative_loss}
\end{equation}
where the first term aims to classify the original node summary embeddings to class 1 and the second term attempt to classify the augmented node summary embeddings to another class 0. With this discriminative pre-text task, encoder $\mathcal{F}$ and projection head $\mathcal{P}$ are trained to extract discriminative features. Besides, this discriminate loss is compatible to batch training techniques on large graphs. 

\textbf{Neural Clustering Module.} This module aims to guide our network to learn clustering-friendly representations. Concretely, based on the learned node representations $\textbf{H}$, cluster center embeddings $\textbf{C} \in \mathbb{R}^{K \times d}$ are initialized in the $K$-Means++ manner \cite{K-Means}, where $K$ denotes the cluster number. It is worth mentioning that the cluster center embeddings are assigned as the optimizable neural parameters with the gradients. Motivated by the dilation and shrink of galaxies in the universe, we design two loss functions to optimize the clustering distribution jointly. 

Firstly, since the universe is expanding, the centers of different galaxies are pushed away from each other. Similarly, we attempt to push away different cluster centers by minimizing the proposed cluster dilation loss as follows. 
\begin{equation} 
\begin{aligned}
\min \mathcal{L}_{\textrm{dilation}} &= \\ \min &\frac{-1}{(K-1)K} \sum_{i=0}^{K-1} \sum_{j=0, j\neq i}^{K-1} \left\|\textbf{C}_i-\textbf{C}_j\right\|_2^2,
\end{aligned}
\label{eq:dilation_loss}
\end{equation}
where $K$ denotes the cluster number. This cluster dilation loss will not bring high time or memory costs even when the sample number $N$ is large. The idea of dilation loss comes from the universe expansion theory \cite{universe_explore}. The cluster centers are like stars with huge masses, and the samples are like planets around the cluster centers. The universe is expanding, and stars are moving apart from each other. Similarly, our cluster dilation loss pushes the cluster centers away from each other.

In addition to universe dilation, the galaxy's center will pull together the planets with gravity. From this observation, a cluster shrink loss is designed to optimize clustering distribution by pulling together samples to cluster centers. Considering the considerable sample number, our shrink loss is compatible with using the mini-batch samples. It is formulated as follows.  
\begin{equation} 
\min \mathcal{L}_{\textrm{shrink}} = \min \frac{1}{BK} \sum_{i=0}^{B-1}\sum_{j=0}^{K-1}\left\|\textbf{H}_i-\textbf{C}_j\right\|_2^2,
\label{eq:shrink_loss}
\end{equation}
where $B$ denotes the batch size. Also, this objective will not bring large time or memory cost since it is linear to batch size $B$ rather than sample number $N$. For this cluster shrink loss, if the clustering algorithm was perfect, we should force the samples close to the nearest cluster center since the perfect clustering algorithm can group the samples with the same ground truth into one cluster. However, for the practical clustering method, this operation easily leads to the confirmation bias problem \cite{Confirmation_bias}. To alleviate this problem, a compromise cluster shrink loss is proposed in Eq. \eqref{eq:shrink_loss}. to guide the samples to be close to all cluster centers. These two intuitive clustering losses optimize the clustering distribution with mini-batch data in shrink and dilation manners, thus endowing scalability and sample discriminative capability of \textit{Dink-Net}.

The overall workflow of our proposed \textit{Dink-Net} is demonstrated in Algorithm \ref{algorithm:dink-net} and the PyTorch-style pseudo-code is given in Appendix.E. Detailed implement about \textit{Dink-Net} can be found in Appendix.C. Next sub-section explores why our method works well on large-scale graphs.

\subsection{Why \textit{Dink-Net} Works Well on Large Graph?}
\label{sec:why_large}
By comparing with the existing methods, this section highlights the advantages of our proposed method from two aspects, including model training and model inference.

\textbf{Model Training.}
As illustrated in Algorithm \ref{algorithm:dink-net}, the training process of \textit{Dink-Net} contains the pre-training and fine-tuning stages. At the pre-train stage, the encoder $\mathcal{F}$ and the projection head $\mathcal{P}$ are optimized by minimizing the discriminate loss $\mathcal{L}_{\text{discri.}}$ in Eq. \eqref{eq:discriminative_loss}. For this loss function, the mini-batch technique can be applied since the discrimination process of each sample is independent. Therefore, given embeddings $\textbf{Z},\textbf{Z}'$, the time complexity and space complexity of calculating $\mathcal{L}_{\text{discri.}}$ is $\mathcal{O}(Bd)$ and $\mathcal{O}(Bd)$, where $B,d$ denote batch size and dimensions of latent features. 

In the fine-tuning procedure, the total loss is formulated below. 
\begin{equation} 
\min \mathcal{L}= \min \left(\mathcal{L}_{\textrm{dilation}}+ \mathcal{L}_{\textrm{shrink}}+\alpha \mathcal{L}_{\textrm{discri.}}\right),
\label{eq:total_loss}
\end{equation}
where $\alpha$ is the trade-off hyper-parameter. We analyze the time and space complexity at this stage as follows. Firstly, given embeddings, the time and space complexity of calculating clustering dilation loss $\mathcal{L}_{\textrm{dilation}}$ in Eq. \eqref{eq:dilation_loss} is $\mathcal{O}(K^2d)$ and $\mathcal{O}(Kd)$, where $K$ denotes the cluster number. Since the $K \ll N$, $\mathcal{L}_{\textrm{dilation}}$ do not expend too much time and space resource even when the sample number grows to a large value. Secondly, the clustering shrink module pulls together the samples to the cluster centers. Considering the large sample space, it optimizes the clustering distribution with mini-batch data rather than operating on all samples. Therefore, in Eq. \eqref{eq:shrink_loss}, it only brings $\mathcal{O}(BKd)$ time complexity and $\mathcal{O}(BK+Bd+Kd)$ space complexity when given the embeddings. Thirdly, calculating $\mathcal{L}_{\textrm{discri.}}$ takes $\mathcal{O}(Bd)$ time complexity and $\mathcal{O}(Bd)$ space complexity. To summarize, at the fine-tune stage, $\mathcal{L}$ takes $\mathcal{O}(BKd+K^2d+Bd)$ time and $\mathcal{O}(BK+Bd+Kd+Kd+Bd) \rightarrow \mathcal{O}(BK+Bd+Kd)$ space costs given the embeddings. Referring to Section \ref{sec:challenge}, it is obvious that our method's time and space costs are much less than that of the existing methods. We attribute this advantage to our proposed cluster dilation and shrink loss functions since they allow our method to optimizing the clustering distribution with mini-batch samples even without performance drops. The experimental evidence can be found in Appendix D.2.

\textbf{Model Inference.}
In the model inference process, with the well-learned cluster center embeddings $\textbf{C} \in \mathbb{R}^{K \times d}$, the assignment of $i$-th sample can be calculated as follows. 
\begin{equation} 
\hat{\textbf{y}}_i = \mathop{\arg\min}_j \|\textbf{H}_i - \textbf{C}_j\|_2,
\label{eq:assignment}
\end{equation}
where $\hat{\textbf{y}} \in \mathbb{R}^N$ denotes the clustering assignment vector. Note that the inference of our method also can be carried out in a mini-batch manner. Therefore, when given embeddings, the time and space complexity of model inference is $\mathcal{O}(BKd)$ and $\mathcal{O}(BK+Bd+Kd)$, where $B$ is batch size. 

The above complexity analyses demonstrate time and space efficiency in theory. Compared with the existing state-of-the-art methods, the main advantages of our method are summarized as follows. 1) \textit{Dink-Net} gets rid of from processing $N \times N$ graph diffusion matrix. 2) Our proposed loss functions allow \textit{Dink-Net} to optimize the clustering distribution with mini-batch data even without performance drops. 3) \textit{Dink-Net} unifies embedding learning and clustering optimization, resulting in clustering-friendly representations. Therefore, this sub-section illustrates that \textit{Dink-Net} can scale well to large-scale graphs in theory. The next section aims to verify the superiority, effectiveness, scalability, and efficiency of \textit{Dink-Net} by extensive experiments.







\section{Experiment}
In this section, we comprehensively evaluate our proposed \textit{Dink-Net} by answering the main questions as follows. 
\begin{itemize}[topsep=4pt, itemsep=3pt]
    \item \textbf{Q1: Superiority.} Does \textit{Dink-Net} outperforms the existing state-of-the-art deep graph clustering methods?
    \item \textbf{Q2: Effectiveness.} Are the proposed node discriminate and neural clustering modules effective? 
    \item \textbf{Q3: Scalability.} Can the proposed method endow the deep graph clustering method scale to large graphs? 
    \item \textbf{Q4: Efficiency.} How about the time and memory efficiency of the proposed method?
    \item \textbf{Q5: Sensitivity.} What is the performance of the proposed method with different hyper-parameters?
    \item \textbf{Q6: Convergence.} Will the proposed loss function, as well as the clustering performance, converge well?
\end{itemize}

\vspace{-5pt}

The answers of \textbf{Q1}-\textbf{Q4} are illustrated in Section 4.2-4.5. In addition, sensitivity analyses (\textbf{Q5}) and convergence analyses (\textbf{Q6}) of \textit{Dink-Net} can be found in Appendix.D.1 and D.2.

\subsection{Experimental Setup}

\subsubsection{Environment}
Experimental results are obtained from the server with four core Intel(R) Xeon(R) Platinum 8358 CPUs @ 2.60GHZ, one NVIDIA A100 GPU (40G), and the PyTorch platform. 

\subsubsection{Dataset}
To evaluate the node clustering performance, we use seven attribute graph datasets, including Cora, CiteSeer, Amazon-Photo, ogbn-arxiv, Reddit, ogbn-products, ogbn-papers100M \cite{ogbn_dataset}. The node numbers of graphs range from $\sim$3~kilo to $\sim$100~million, and the edge numbers of graphs range from $\sim$5~kilo to $\sim$1~billion. The statistical information is summarized in Table~2 of Appendix.

\subsubsection{Evaluation Protocol}
To evaluate the clustering methods, the predicted clustering assignment vector is firstly mapped to the ground truth by the Kuhn-Munkres algorithm \cite{Kuhn-Munkres}. Then, the clustering performance is evaluated by four widely-used metrics \cite{DCRN}, including accuracy (ACC), normalized mutual information (NMI), adjusted rand index (ARI), and F1-score (F1). All results are obtained under three runs with different random seeds.

\subsubsection{Compared Baseline}
To demonstrate the superiority of the proposed method, we conduct comprehensive experiments to compare our \textit{Dink-Net} with a variety of baseline methods. Concretely, the classical clustering method $K$-Means \cite{K-Means} uses the idea of exception maximum to separate samples. Additionally, the deep clustering methods \cite{DEC,IDEC,DCN,AdaGAE} apply the deep neural networks to assist clustering. Moreover, the deep graph clustering methods \cite{node2vec,DGI,AGE,mincutpool,MVGRL,BYOL_graph,GRACE,AGC-DRR,DCRN,S3GC,MGAE,DAEGC,ARGA,SDCN,GDCL,DFCN,AFGRL} utilize graph neural networks to reveal graph structure and then group nodes into different clusters. 

\begin{table*}[!t]
\centering
\setlength{\tabcolsep}{4pt}
\resizebox{\linewidth}{!}{
\begin{tabular}{@{}ccccccccccccccccc@{}}
\toprule
\multirow{2}{*}{\textbf{Dataset}}         & \multirow{2}{*}{\textbf{Metric}} & \multirow{2}{*}{\textbf{K-Means}} & \multirow{2}{*}{\textbf{DEC}} & \multirow{2}{*}{\textbf{DCN}} & \multirow{2}{*}{\textbf{node2vec}} & \multirow{2}{*}{\textbf{DGI}} & \multirow{2}{*}{\textbf{AGE}} & \multirow{2}{*}{\textbf{MinCutPool}} & \multirow{2}{*}{\textbf{MVGRL}} & \multirow{2}{*}{\textbf{BGRL}} & \multirow{2}{*}{\textbf{GRACE}} & \multirow{2}{*}{\textbf{ProGCL}} & \multirow{2}{*}{\textbf{AGC-DRR}} & \multirow{2}{*}{\textbf{DCRN}} & \multirow{2}{*}{\textbf{S$^3$GC}} & \multirow{2}{*}{\textbf{Ours}} \\
&                                  &                                  &                               &                               &                                    &                               &                               &                                      &                                 &                                &                                 &                                  &                                   &                                &                                &                                \\ \midrule
\multirow{4}{*}{\textbf{Cora}}            & \textbf{ACC}                     & 33.80                            & 46.50                         & 49.38                         & 61.20                              & 72.60                         & 73.50                         & 49.00                                & {\ul 76.30}                     & 74.20                          & 73.90                           & 57.13                            & 40.62                             & 61.93                          & 74.20                          & \textbf{78.10}                 \\
& \textbf{NMI}                     & 14.98                            & 23.54                         & 25.65                         & 44.40                              & 57.10                         & 57.58                         & 41.00                                & {\ul 60.80}                     & 58.40                          & 57.00                           & 41.02                            & 18.74                             & 45.13                          & 58.80                          & \textbf{62.28}                 \\
& \textbf{ARI}                     & 8.60                             & 15.13                         & 21.63                         & 32.90                              & 51.10                         & 50.10                         & 30.80                                & {\ul 56.60}                     & 53.40                          & 52.70                           & 30.71                            & 14.80                             & 33.15                          & 54.40                          & \textbf{61.61}                 \\
& \textbf{F1}                      & 30.26                            & 39.23                         & 43.71                         & 62.10                              & 69.20                         & 69.28                         & 51.80                                & 71.60                           & 69.10                          & {\ul 72.50}                     & 45.68                            & 31.23                             & 49.50                          & 72.10                          & \textbf{72.66}                 \\ \midrule
\multirow{4}{*}{\textbf{CiteSeer}}        & \textbf{ACC}                     & 39.32                            & 55.89                         & 57.08                         & 42.10                              & 68.60                         & 69.73                         & 53.70                                & 62.83                           & 67.50                          & 63.10                           & 65.92                            & 68.32                             & {\ul 69.86}                    & 68.80                          & \textbf{70.36}                 \\
& \textbf{NMI}                     & 16.94                            & 28.34                         & 27.64                         & 24.00                              & 43.50                         & {\ul 44.93}                   & 29.50                                & 40.69                           & 42.20                          & 39.90                           & 39.59                            & 43.28                             & 44.86                          & 44.10                          & \textbf{45.87}                 \\
& \textbf{ARI}                     & 13.43                            & 28.12                         & 29.31                         & 11.60                              & 44.50                         & 45.31                         & 26.20                                & 34.18                           & 42.80                          & 37.70                           & 36.16                            & 45.34                             & {\ul 45.64}                    & 44.80                          & \textbf{46.96}                 \\
& \textbf{F1}                      & 36.08                            & 52.62                         & 53.80                         & 40.10                              & 64.30                         & 64.45                         & 51.60                                & 59.54                           & 63.10                          & 60.30                           & 57.89                            & 64.82                             & {\ul 64.83}                    & 64.30                          & \textbf{65.96}                 \\ \midrule
\multirow{4}{*}{\makecell[c]{\textbf{Amazon-} \\ \textbf{Photo}}}    & \textbf{ACC}                     & 27.22                            & 47.22                         & 48.25                         & 27.58                              & 43.03                         & 75.98                         & 54.67                                & 41.07                           & 66.54                          & 67.66                           & 51.53                            & 76.81                             & {\ul 79.94}                    & 75.15                          & \textbf{81.71}                 \\
& \textbf{NMI}                     & 13.23                            & 37.35                         & 38.76                         & 11.53                              & 33.67                         & 65.38                         & 50.02                                & 30.28                           & 60.11                          & 53.46                           & 39.56                            & 66.54                             & {\ul 73.70}                    & 59.78                          & \textbf{74.36}                 \\
& \textbf{ARI}                     & 5.50                             & 18.59                         & 20.80                         & 4.92                               & 22.15                         & 55.89                         & 34.43                                & 18.77                           & 44.14                          & 42.74                           & 34.18                            & 60.15                             & {\ul 63.69}                    & 56.13                          & \textbf{68.40}                 \\
& \textbf{F1}                      & 23.96                            & 46.71                         & 47.87                         & 21.52                              & 35.17                         & 71.74                         & 53.02                                & 32.88                           & 63.08                          & 60.30                           & 31.97                            & 71.03                             & {\ul 73.82}                    & 72.85                          & \textbf{73.92}                 \\ \midrule
\multirow{4}{*}{\textbf{ogbn-arXiv}}      & \textbf{ACC}                     & 18.11                            & 21.25                         & 19.91                         & 29.00                              & 31.40                         & \multirow{4}{*}{OOM}          & 24.20                                & \multirow{4}{*}{OOM}            & 22.70                          & \multirow{4}{*}{OOM}            & 29.86                            & \multirow{4}{*}{OOM}              & \multirow{4}{*}{OOM}           & {\ul 35.00}                    & \textbf{43.68}                 \\
& \textbf{NMI}                     & 22.13                            & 25.14                         & 23.81                         & 40.60                              & 41.20                         &                               & 38.00                                &                                 & 32.10                          &                                 & 37.51                            &                                   &                                & \textbf{46.30}                 & {\ul 43.73}                    \\
& \textbf{ARI}                     & 7.43                             & 10.28                         & 8.25                          & 19.00                              & 22.30                         &                               & 13.90                                &                                 & 13.00                          &                                 & 25.74                            &                                   &                                & {\ul 27.00}                    & \textbf{35.22}                 \\
& \textbf{F1}                      & 12.94                            & 15.57                         & 13.06                         & 22.00                              & 23.00                         &                               & 19.80                                &                                 & 16.60                          &                                 & 21.79                            &                                   &                                & {\ul 23.00}                    & \textbf{26.92}                 \\ \midrule
\multirow{4}{*}{\makecell[c]{\textbf{ogbn-} \\ \textbf{products}}}   & \textbf{ACC}                     & 18.11                            & 23.79                         & 24.50                         & 35.70                              & 32.00                         & \multirow{4}{*}{OOM}          & 25.70                                & \multirow{4}{*}{OOM}            & \multirow{4}{*}{OOM}           & \multirow{4}{*}{OOM}            & 35.21                            & \multirow{4}{*}{OOM}              & \multirow{4}{*}{OOM}           & {\ul 40.20}                    & \textbf{41.09}                 \\
& \textbf{NMI}                     & 22.13                            & 24.47                         & 21.92                         & 48.90                              & 46.70                         &                               & 43.00                                &                                 &                                &                                 & 46.59                            &                                   &                                & \textbf{53.60}                 & {\ul 50.78}                    \\
& \textbf{ARI}                     & 7.43                             & 9.05                          & 10.96                         & 17.00                              & 17.40                         &                               & 13.00                                &                                 &                                &                                 & 19.87                            &                                   &                                & \textbf{23.00}                 & {\ul 21.08}                    \\
& \textbf{F1}                      & 12.94                            & 13.54                         & 13.95                         & 24.70                              & 19.20                         &                               & 18.00                                &                                 &                                &                                 & 21.55                            &                                   &                                & {\ul 25.00}                    & \textbf{25.15}                 \\ \midrule
\multirow{4}{*}{\textbf{Reddit}}          & \textbf{ACC}                     & 8.90                             & \multirow{4}{*}{OOM}          & \multirow{4}{*}{OOM}          & 70.90                              & 22.40                         & \multirow{4}{*}{OOM}          & \multirow{4}{*}{-}                   & \multirow{4}{*}{OOM}            & \multirow{4}{*}{OOM}           & \multirow{4}{*}{OOM}            & 65.41                            & \multirow{4}{*}{OOM}              & \multirow{4}{*}{OOM}           & {\ul 73.60}                    & \textbf{76.03}                 \\
& \textbf{NMI}                     & 11.40                            &                               &                               & 79.20                              & 30.60                         &                               &                                      &                                 &                                &                                 & 70.48                            &                                   &                                & \textbf{80.70}                 & {\ul 78.91}                    \\
& \textbf{ARI}                     & 2.90                             &                               &                               & 64.00                              & 17.00                         &                               &                                      &                                 &                                &                                 & 63.42                            &                                   &                                & \textbf{74.50}                 & {\ul 71.34}                    \\
& \textbf{F1}                      & 6.80                             &                               &                               & 55.10                              & 18.30                         &                               &                                      &                                 &                                &                                 & 51.45                            &                                   &                                & {\ul 56.00}                    & \textbf{67.95}                 \\ \midrule
\multirow{4}{*}{\makecell[c]{\textbf{ogbn-} \\ \textbf{papers100M}}} & \textbf{ACC}                     & 14.60                            & \multirow{4}{*}{OOM}          & \multirow{4}{*}{OOM}          & {\ul 17.50}                        & 15.10                         & \multirow{4}{*}{OOM}          & \multirow{4}{*}{OOM}                 & \multirow{4}{*}{OOM}            & \multirow{4}{*}{OOM}           & \multirow{4}{*}{OOM}            & \multirow{4}{*}{OOM}             & \multirow{4}{*}{OOM}              & \multirow{4}{*}{OOM}           & 17.30                          & \textbf{26.67}                 \\
& \textbf{NMI}                     & 37.33                            &                               &                               & 38.00                              & 41.60                         &                               &                                      &                                 &                                &                                 &                                  &                                   &                                & {\ul 45.30}                    & \textbf{54.92}                 \\
& \textbf{ARI}                     & 7.54                             &                               &                               & {\ul 11.20}                        & 9.60                          &                               &                                      &                                 &                                &                                 &                                  &                                   &                                & 11.00                          & \textbf{18.01}                 \\
& \textbf{F1}                      & 10.45                            &                               &                               & 11.10                              & 11.10                         &                               &                                      &                                 &                                &                                 &                                  &                                   &                                & {\ul 11.80}                    & \textbf{19.48}                 \\ 
\bottomrule
\end{tabular}}
\vspace{-5pt}
\caption{Clustering performance ($\%$) of our method and fourteen state-of-the-art baselines. The bold and underlined values are the best and the runner-up results. ``OOM'' indicates that the method raises the out-of-memory failure. ``-'' denotes that the methods do not converge.}
\label{compare_table}
\end{table*}

\begin{figure*}[!t]
\centering
\small
\begin{minipage}{0.139\linewidth}
\centerline{\includegraphics[width=\textwidth]{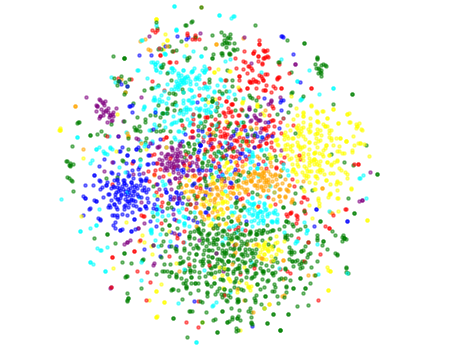}}
\vspace{2pt}
\centerline{Raw Attribute}
\end{minipage}
\begin{minipage}{0.139\linewidth}
\centerline{\includegraphics[width=\textwidth]{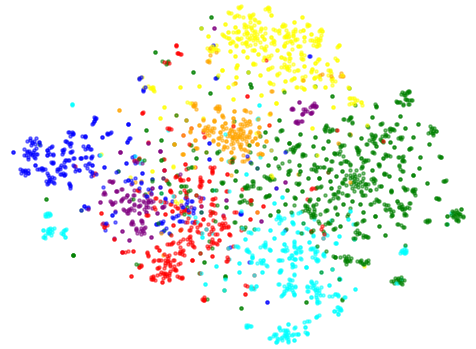}}
\vspace{2pt}
\centerline{DEC}
\end{minipage}
\begin{minipage}{0.139\linewidth}
\centerline{\includegraphics[width=\textwidth]{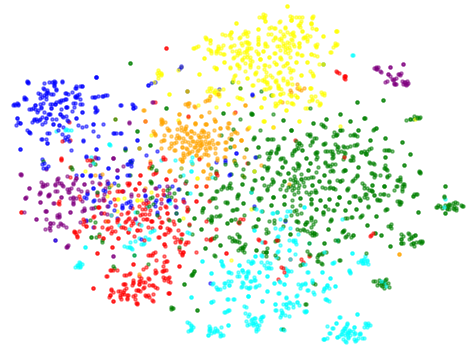}}
\vspace{2pt}
\centerline{MVGRL}
\end{minipage}
\begin{minipage}{0.139\linewidth}
\centerline{\includegraphics[width=\textwidth]{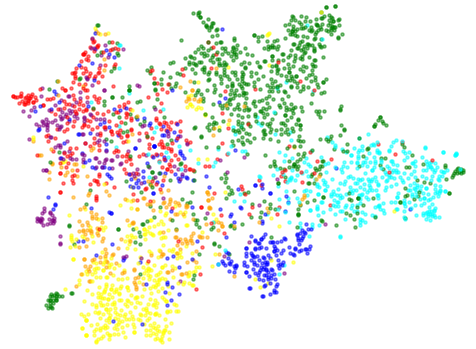}}
\vspace{2pt}
\centerline{MinCutPool}
\end{minipage}
\begin{minipage}{0.139\linewidth}
\centerline{\includegraphics[width=\textwidth]{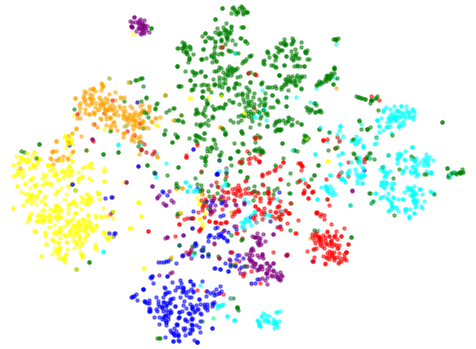}}
\vspace{2pt}
\centerline{DCRN}
\end{minipage}
\begin{minipage}{0.139\linewidth}
\centerline{\includegraphics[width=\textwidth]{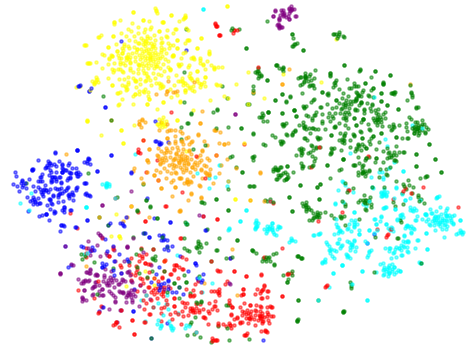}}
\vspace{2pt}
\centerline{S$^3$GC}
\end{minipage}
\begin{minipage}{0.139\linewidth}
\centerline{\includegraphics[width=\textwidth]{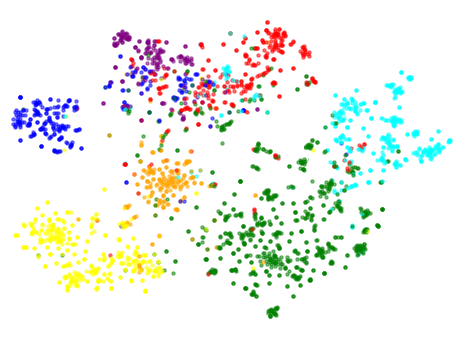}}
\vspace{2pt}
\centerline{Ours}
\end{minipage}
\caption{\textit{t}-SNE visualization of seven methods on the Cora dataset.}
\label{fig:visualization}  
\end{figure*}

\subsection{Superiority}
\label{sec:performance}
This section answers the question \textbf{Q1}. To illustrate the superiority of the proposed method, extensive experiments are carried out to compare \textit{Dink-Net} with the existing state-of-the-art methods. Four conclusions are demonstrated by carefully analyzing the results in Table~\ref{compare_table}. 

1) The performance of the traditional method $K$-Means is limited. For example, on the Reddit dataset, $K$-Means merely achieves 8.90\% ACC. The main reason is that it lacks representation learning, leading to indiscriminate samples.

2) The deep clustering methods DEC \cite{DEC} and DCN \cite{DCN} achieve an un-promising performance because they merely extract features from node attributes but ignore the structural information. For example, on Cora dataset, our method \textit{Dink-Net} outperforms DEC by about 38.74\% NMI.

3) For the deep graph clustering method, node2vec \cite{node2vec} merely takes care of the graph structure and overlooks the node attributes. Besides, the graph representation learning methods \cite{DGI,AGE,MVGRL,BYOL_graph,GRACE,ProGCL} can not optimize embedding and clustering in a unified framework. Therefore, these methods gain the sub-optimal clustering performance compared to our proposed method. For example, on the CiteSeer dataset, compared with BGRL, our method achieves about 2.86\% ACC improvement. 

4) Our proposed method outperforms the recent state-of-the-art deep graph clustering methods. For example, on the ogbn-papers100M dataset, \textit{Dink-Net} achieves $9.62\%$ NMI increment compared to the runner-up method S$^3$GC \cite{S3GC}. The reason contains two aspects as follows. Firstly, the discriminate pre-text task in the representation learning process enhances the discriminative capability of samples. Secondly, the clustering modules unify the representation learning and the clustering process, guiding models to learn clustering-friendly representations. 


Moreover, in order to intuitively demonstrate the superiority of our proposed \textit{Dink-Net}, we visualize the learned node representations via the \textit{t}-SNE algorithm \cite{T_SNE}. As shown in Figure \ref{fig:visualization}, we find that our proposed method better reveals cluster structure in the latent space. Due to the page limitation, the additional experimental results and analyses are presented in Appendix D.4.

\vspace{-3pt}

\subsection{Effectiveness}
\label{sec:effectiveness}
\begin{figure}[!t]
\centering
\small
\begin{minipage}{0.49\linewidth}
\centerline{\includegraphics[width=1\textwidth]{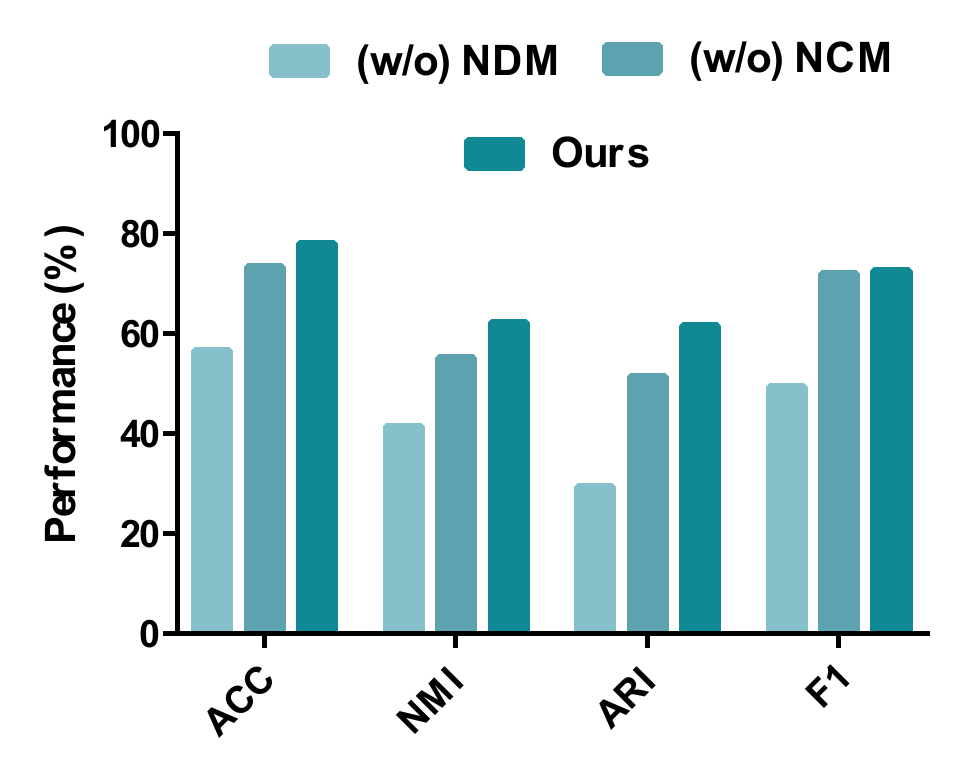}}
\centerline{(a) Cora}
\vspace{2pt}
\end{minipage}
\begin{minipage}{0.49\linewidth}
\centerline{\includegraphics[width=1\textwidth]{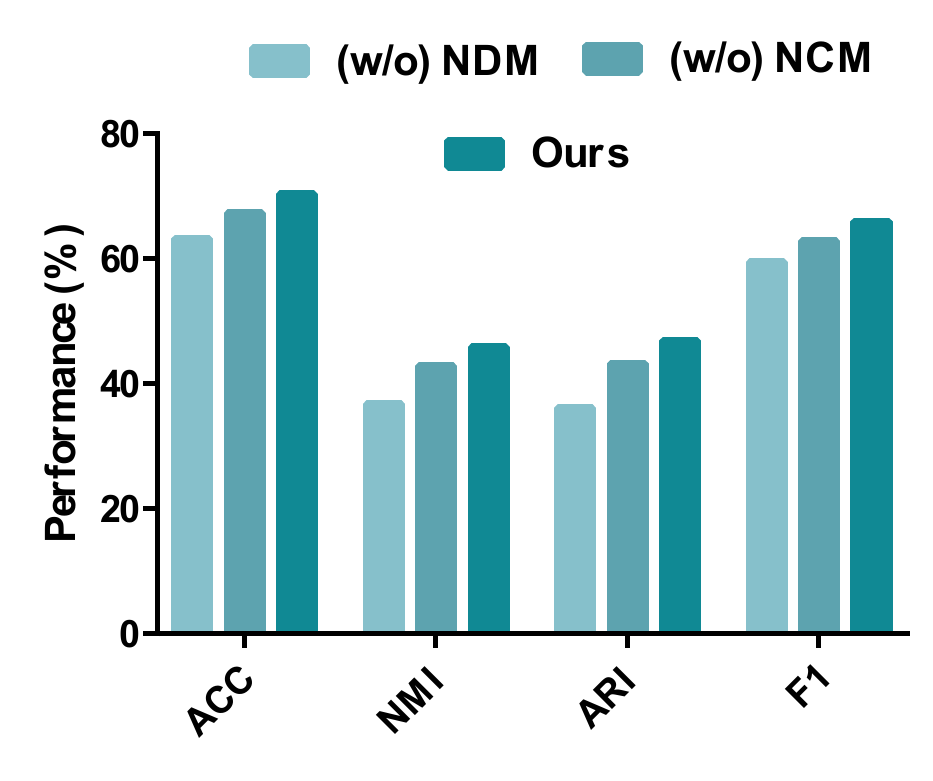}}
\centerline{(b) CiteSeer}
\vspace{2pt}
\end{minipage}
\begin{minipage}{0.49\linewidth}
\centerline{\includegraphics[width=1\textwidth]{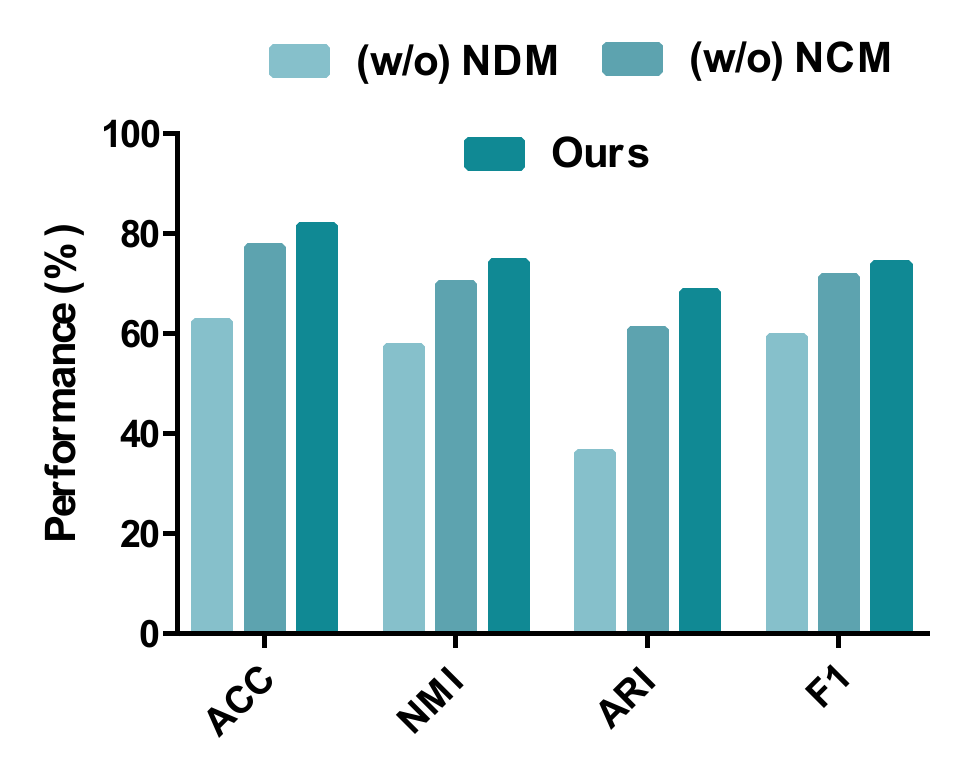}}
\centerline{(c) Amazon-Photo}
\vspace{2pt}
\end{minipage}
\begin{minipage}{0.49\linewidth}
\centerline{\includegraphics[width=1\textwidth]{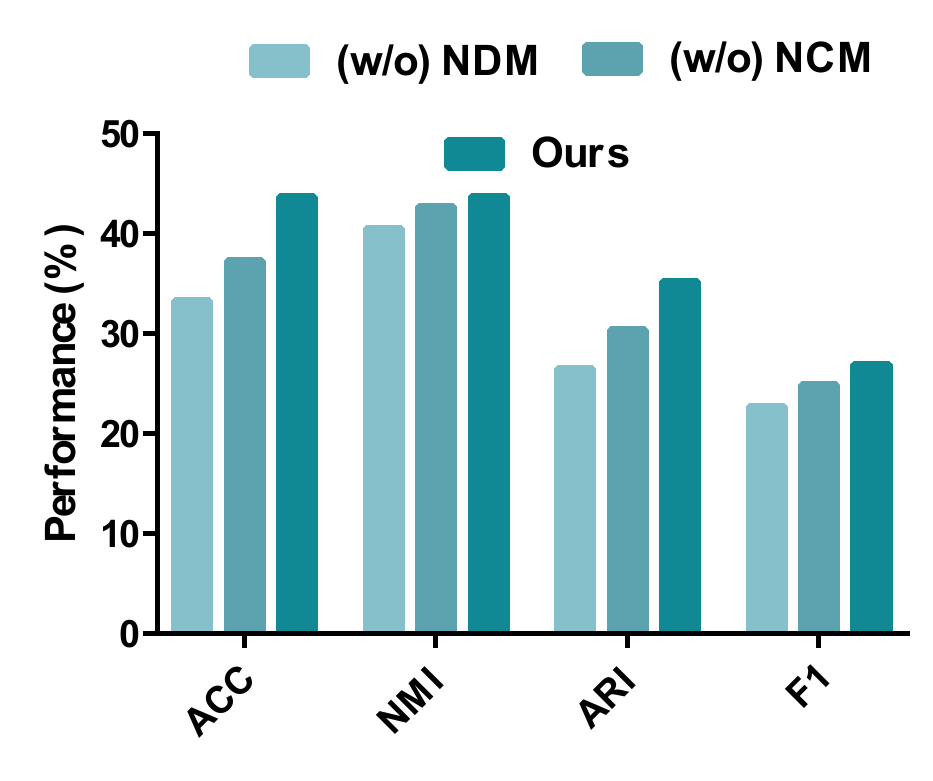}}
\centerline{(d) ogbn-arXiv}
\vspace{2pt}
\end{minipage}
\caption{Ablation studies of the proposed modules on four datasets. ``(w/o) NDM'' denotes \textit{Dink-Net} without the node discriminate module. ``(w/o) NDM'' denotes \textit{Dink-Net} without the neural clustering module. ``Ours'' denotes our proposed \textit{Dink-Net}.}
\label{ablation_study}
\end{figure}

The question \textbf{Q2} is answered in this section. To verify the effectiveness of the proposed modules, we carefully conduct ablation studies on four datasets. Specifically, as shown in Figure \ref{ablation_study}, our method is denoted as ``Ours''. Additionally, our method without the node discriminate module and without the neural clustering module is denoted as ``(w/o) NDM'' and ``(w/o) NCM'', respectively. From the experimental results in Figure \ref{ablation_study}, three observations are presented as follows. 

\vspace{-5pt}

1) ``(w/o) NDM'' can not achieve expected clustering performance since it lacks the strong representation learning capability of the node discriminate module. 

2) The results indicate that ``(w/o) NCM'' becomes the runner-up. Although the node discriminate module endows strong representation capability, it can not learn clustering-friendly features since the process of embedding, and clustering is detached. 

3) Our \textit{Dink-Net} achieves the best clustering performance because it unifies representation learning and clustering optimization to extract clustering-friendly features. 



    

The above results and observations prove the effectiveness of the node discriminate and neural clustering module in \textit{Dink-Net}. Both of them can boost performance.

\begin{table*}[!t]
\centering
\resizebox{\linewidth}{!}{
\begin{tabular}{@{}ccccc@{}}
\toprule
\textbf{Method}     & \textbf{Time Complexity (per iteration)} & \textbf{Space Complexity} &  \textbf{\makecell[c]{Time Cost (s)} } & \textbf{\makecell[c]{Memory Cost (MB)} }\\ \midrule
\textbf{DGI}                 &        $\mathcal{O}(ED+Nd^2)$                    &         $\mathcal{O}(E+Nd+d^2)$                   &             19.03         &    3798                \\
\textbf{MVGRL}               &  $\mathcal{O}(N^2d+Nd^2)$                        &          $\mathcal{O}(N^2+Nd+d^2)$                 &                   168.20   &       9466             \\
\textbf{S$^3$GC}                &         $\mathcal{O}(NSd^2)$                 &              $\mathcal{O}(Nd+BSd+d^2)$             &                  508.21    &           1474         \\ 
\textbf{\textit{Dink-Net (Ours)}}                &         $\mathcal{O}(BKd+K^2d+Bd)$                 &              $\mathcal{O}(BK+Bd+Kd)$             &             35.09         &        1248            \\ \bottomrule
\end{tabular}}
\caption{Time and space analyses of various methods. The experimental GPU memory costs and time costs are obtained on Cora dataset.}
\label{table:complexity}
\end{table*}

\subsection{Scalability}
\label{sec:scalability}
This section attempts to answer the question \textbf{Q3}. To verify the scalability of the proposed method, we conduct experiments on seven graph datasets with different scales. For instance, the Cora dataset contains 2708 nodes, and ogbn-papers100M contains $\sim$111~million nodes. The statistics of these datasets can be found in Table~2 in Appendix. Table~\ref{compare_table} provides the clustering performance on these datasets. From these results, we have four observations as follows. 

1) The traditional clustering method $K$-Means can complete the clustering process on seven datasets. However, there are two drawbacks as follows. Firstly, it fails to achieve promising performance since it lacks the representation learning process. For example, on the Reddit dataset our method outperforms $K$-Means $67.13\%$ from the aspect of ACC. Secondly, it takes a long running time on large-scale graph datasets, e.g., $\sim$5~days for papers100M dataset on CPU.

2) The deep clustering methods raise the out-of-memory failure on the Reddit and ogbn-papers100M datasets. The main reason for enormous memory costs is that the KL divergence loss \cite{DEC,IDEC} estimates and sharpens the cluster distribution with all samples, leading to enormous memory costs. Also, they neglect the graph structure leading to worse performance.

3) The most of deep graph clustering methods easily lead to the out-of-memory problem because some methods \cite{MVGRL,DCRN,AGC-DRR} need to process $N \times N$ dense graph diffusion matrix, which is inefficient on time and memory. For the scalable methods like node2vec \cite{node2vec}, DGI \cite{DGI}, and S$^3$GC \cite{S3GC}, they separate the embedding and clustering, leading to sub-optimal performance.

4) Our proposed \textit{Dink-Net} scales well to all seven datasets and achieves promising performances. The reasons and analyses are demonstrated in Section \ref{sec:why_large}.

    




Through the above observations, we conclude that the existing methods easily lead to the out-of-memory problem or the long ruining time problem. But our method can endow the models with excellent scalability to large-scale graphs.

    

    


\subsection{Efficiency}
\label{sec:efficiency}
This section attempts to answer the question \textbf{Q4}. To verify the efficiency of \textit{Dink-Net}, we test the time and memory costs of various methods on the Cora and ogbn-papers100M datasets. The main observations (See Table~\ref{compare_table} and Table~\ref{table:complexity} ) and analyses are illustrated. 

1) For the time costs, on ogbn-papers100M dataset, the scalable baselines node2vec\cite{node2vec}, DGI\cite{DGI}, and S$^3$GC\cite{S3GC} run in $\sim$24 hours. Differently, our proposed method takes $\sim$9~hours for model pre-training and $\sim$3~hours for model fine-tuning. From the theoretical view, at the pre-training stage, the time complexity of calculating discriminative loss is $\mathcal{O}(Bd)$, which is linear to batch size. Moreover, at fine-tuning state, calculating the clustering loss takes $\mathcal{O}(BKd+K^2d+Kd)$ time complexity, which is also linear to batch size.

2) For the memory costs on ogbn-papers100M dataset, most baseline methods raise the out-of-memory problem on 40GB GPU. But our method takes $\sim$20GB~GPU~memory during training. In addition, from a theoretical perspective, calculating discriminative loss and clustering loss take $\mathcal{O}(Bd)$ and $\mathcal{O}(BK+Bd+Kd)$ space complexity, which are both linear to node number in a mini-batch.

    

    


These results and analyses demonstrate the efficiency of our proposed method in both the time and memory aspects. Due to page limitation, the additional experiments and complexity analyses can be found in Appendix.B.


\section{Conclusion}
This work aims to scale deep graph clustering to large graphs. It begins with analyzing the drawbacks of existing methods. Firstly, part of the method must process a space-consuming graph diffusion matrix. Secondly, some algorithms must optimize clustering distribution with all nodes, easily resulting in the out-of-memory problem. Thirdly, the scalable S$^3$GC achieves sub-optimal performance since it separates representation and clustering. To solve this problem, a novel scalable deep graph clustering termed $\textit{Dink-Net}$ is proposed under the guidance idea of dilation and shrink. Our method contains the node discriminate and neural clustering module. With these designs, we unify representation learning and clustering optimization into an end-to-end framework, guiding network to learn clustering-friendly features. Also, cluster dilation and shrink loss functions allow our method to optimize clustering distribution with mini-batch data. Extensive experiments and theoretical analyses verify the superiority. In the future, it is worth to extending $\textit{Dink-Net}$ to heterogeneous graphs \cite{zheng2021heterogeneous}, heterophily graphs\cite{liu2022beyond}, knowledge graphs \cite{liang2022relational}, and molecular graphs \cite{mole-bert}.

\section{Acknowledgments}

We thank all anonymous reviewers and program chairs for their constructive and helpful reviews. This work was supported by the National Key R\&D Program of China (project no. 2020AAA0107100) and the National Natural Science Foundation of China (project no. 62276271). Besides, this work was also supported by the National Key R\&D Program of China (Project 2022ZD0115100), the National Natural Science Foundation of China (Project U21A20427), the Research Center for Industries of the Future (Project WU2022C043), and the Competitive Research Fund (Project WU2022A009) from the Westlake Center for Synthetic Biology and Integrated Bioengineering.

\bibliography{2_ref}
\bibliographystyle{icml2023}

\end{document}


\setcounter{page}{13}
\icmltitle{Appendix of ``Dink-Net: Neural Clustering on Large Graph''}

\appendix

\section{Notations \& Datasets}
The basic notations are summarized in Table \ref{notation_table}. Table \ref{statistics_information} lists the statistical information about seven datasets. These datasets have various scales. For example, CiteSeer has about 3.3K nodes and 4.6K edges, while ogbn-papers100M has about 111M nodes and 1.6B edges. In addition, the network densities of these graphs are various. Concretely, the density of Cora is 0.07\% while the density of Amazon-Photo is 0.25\%.

\begin{table}[h]
\centering
\scalebox{1.0}{
\begin{tabular}{@{}cc@{}}
\toprule
\textbf{Notation} & \textbf{Meaning}  \\ 
\hline   
$\boldsymbol{G}$       &    Attribute Graph                
\\
$N$       &    Sample Number
\\
$D$       &    Attribute Dimension Number
\\
$d$       &    Latent Feature Dimension Number
\\
$\mathcal{F}$     &  Encoding Network
\\
$\mathcal{C}$     &  Clustering Method
\\
$\textbf{X} \in \mathbb{R}^{N \times D}$     &  Attribute Matrix       \\
$\textbf{A} \in \mathbb{R}^{N \times N}$    &     Adjacency Matrix     \\  
$\textbf{H} \in \mathbb{R}^{N \times d}$    &   Node Embedding Matrix  \\
$\textbf{g} \in \mathbb{R}^{N \times 1}$    &   Node Summary Vector  \\
$\textbf{C} \in \mathbb{R}^{K \times d}$    &   Cluster Center Embedding Matrix  \\
$\hat{\textbf{y}} \in \mathbb{R}^{N \times 1}$  &  Clustering Assignment Vector  \\
$\textbf{y} \in \mathbb{R}^{N \times 1}$  &  Sample Label Vector  \\
\bottomrule
\end{tabular}}
\caption{Basic Notations}
\label{notation_table}
\end{table}

\begin{table*}[h]
\centering
\scalebox{1.0}{
\begin{tabular}{@{}cccccc@{}}
\toprule
\textbf{Dataset} & \textbf{Type}    & \textbf{\# Nodes} & \textbf{\# Edges} & \textbf{\# Feature Dims} & \textbf{\# Classes}  \\ 
\midrule
Cora             & Attributed Graph & 2,708             & 5,278             & 1,433                    & 7                    \\
CiteSeer         & Attributed Graph & 3,327             & 4,614             & 3,703                    & 6                    \\
Amazon-Photo             & Attributed Graph & 7,650             & 119,081           & 745                      & 8                    \\
ogbn-arxiv       & Attributed Graph & 169,343           & 1,166,243         & 128                      & 40                   \\
Reddit           & Attributed Graph & 232,965           & 23,213,838        & 602                      & 41                   \\
ogbn-products    & Attributed Graph & 2,449,029         & 61,859,140        & 100                      & 47                   \\
ogbn-papers100M  & Attributed Graph & 111,059,956       & 1,615,685,872     & 128                      & 172                  \\
\bottomrule
\end{tabular}}
\caption{The statistical information of seven datasets.}
\label{statistics_information}
\end{table*}

\section{Time and Space Analyses}
In this section, we analyze and summarize the time and space complexity of the various baseline methods, including spectral clustering \cite{spectral_clustering}, K-Means \cite{K-Means}, DEC \cite{DEC}, node2vec \cite{node2vec}, DGI \cite{DGI}, MVGRL \cite{MVGRL}, GRACE \cite{GRACE}, BGRL \cite{BYOL_graph}, S$^3$GC \cite{S3GC}, and our proposed \textit{Dink-Net} in Table \ref{table:complexity}. Here, $N$ denotes the node number in the graph, $B$ denotes the batch size, $K$ denotes the cluster number, $E$ denotes the edges number of the graph, $S$ denotes the average degree of the graph, $D$ denotes the dimensions of node attributes, and $d$ denotes the dimensions of latent features. From these analyses, we find that the complexity of most methods will become unacceptable when the sample number reaches a tremendous value. Different, our method's time and memory complexity is linear to the batch size, alleviating the out-of-memory and long-running time problems. In addition, we also test the memory and time costs of these methods via experiments on the Cora dataset. From these experimental results, we find that our proposed \textit{Dink-Net} also achieves efficient results even without batch-training techniques on a small dataset. More importantly, \textit{Dink-Net} is compatible with the mini-batch training technique even without performance drops. Therefore it scales well with the large graphs. Experimental evidence can be found in Figure \ref{fig:Sensitivity}.



\begin{table}[h]
\centering
\scalebox{0.90}{
\begin{tabular}{@{}ccccc@{}}
\toprule
\textbf{Method}     & \textbf{Time Complexity (per iteration)} & \textbf{Space Complexity} & \textbf{GPU Memory Cost (MB)} & \textbf{Time Cost (s)} \\ \midrule
\textbf{Spectral Clustering} &       $\mathcal{O}(N^3)$                   &    $\mathcal{O}(N^2)$                       &      -                &       29.31             \\
\textbf{K-Means}             &     $\mathcal{O}(NKD)$                     &           $\mathcal{O}(NK+ND+KD)$                &        -              &          6.01          \\
\textbf{DEC}            &        $\mathcal{O}(NKd)$                  &             $\mathcal{O}(NK+Nd+Kd)$              &          1294            &       14.59             \\
\textbf{node2vec}            &        $\mathcal{O}(Bd)$                  &             $\mathcal{O}(Nd)$              & -                     &         111.03           \\
\textbf{DGI}                 &        $\mathcal{O}(ED+Nd^2)$                    &         $\mathcal{O}(E+Nd+d^2)$                   &            3798          &               19.03     \\
\textbf{MVGRL}               &  $\mathcal{O}(N^2d+Nd^2)$                        &          $\mathcal{O}(N^2+Nd+d^2)$                 &                  9466    &           168.20         \\
\textbf{GRACE}               &     $\mathcal{O}(N^2d+Ed+d^2)$                      &         $\mathcal{O}(E+Nd)$                  &                 1292     &          44.77          \\
\textbf{BGRL}                &        $\mathcal{O}(Ed+Nd^2)$                  &         $\mathcal{O}(E+Nd+d^2)$                  &                1258      &       44.18             \\
\textbf{S$^3$GC}                &         $\mathcal{O}(NSd^2)$                 &              $\mathcal{O}(Nd+BSd+d^2)$             &                   1474   &      508.21              \\ 
\textbf{\textit{Dink-Net}}                &         $\mathcal{O}(BKd+K^2d+Bd)$                 &              $\mathcal{O}(BK+Bd+Kd)$             &             1248         &            35.09        \\ \bottomrule
\end{tabular}}
\caption{Time and space analyses of various methods. The experimental costs are obtained on the Cora dataset. ``-'' means ruining on CPU.}
\label{table:complexity}
\end{table}

\section{Design Details \& Hyper-parameter Settings}
In this section, we introduce the design details of our proposed method and summarize the hyper-parameter settings. Following the existing works \cite{GGD,DGI}, for the encoder $\mathcal{F}$, we adopt the graph convolutional network (GCN) \cite{GCN}. Besides, we use multilayer perceptron (MLP) \cite{MLP} as the projector in \textit{Dink-Net}. Next, we report the hyper-parameter settings of our method in Table \ref{table:hyper-parameter}. Here, $T$ is the epoch number of pre-training, $T'$ is the epoch number of fine-tuning, $\beta$ is the learning rate of pre-training, $\beta'$ is the learning rate of fine-tuning, $\alpha$ is the trade-off parameter, $B$ is the batch size, and $d$ is the dimension number of latent features.

\begin{table}[h]
\centering
\scalebox{1.0}{
\begin{tabular}{@{}cccccccc@{}}
\toprule
\textbf{}                & $T$ & $\beta$ & $T'$ & $\beta'$ & $\alpha$ & $B$ & $d$ \\ \midrule
\textbf{Cora}            & 200                     & 1e-3             & 200                     & 1e-2              & 1e-10       & -    & 512                  \\
\textbf{CiteSeer}        & 100                     & 5e-4             & 200                     & 1e-2              & 1e-10       & -    & 1536                 \\
\textbf{Amazon-Photo}    & 2000                    & 5e-4             & 100                     & 1e-2              & 1e-10       & -    & 512                  \\
\textbf{ogbn-arXiv}      & 1                       & 1e-4             & 100                     & 1e-4              & 1e-10       & 8192                & 1500                 \\
\textbf{ogbn-products}   & 10                      & 1e-3             & 10                      & 1e-2              & 1e-10       & 8192                & 1024                 \\
\textbf{Reddit}          & 10                       & 1e-4             & 1                      & 1e-5              & 1e-10       & 10240                & 512                  \\
\textbf{ogbn-papers100M} & 1                       & 1e-4             & 1                      & 1e-5              & 1e-10       & 10240               & 256                  \\ \bottomrule
\end{tabular}}
\caption{Hyper-parameter settings of our proposed method. ``-'' denotes that it does not use mini-batch training.}
\label{table:hyper-parameter}
\end{table}

\section{Additional Experimental Result}

\subsection{Compare Experiment}

\begin{table}[!t]
\centering
\scalebox{0.9}{
\begin{tabular}{@{}cccccccccccc@{}}
\toprule
\multirow{2}{*}{\textbf{Dataset}}         & \multirow{2}{*}{\textbf{Metric}} & \multirow{2}{*}{\textbf{IDEC}} & \multirow{2}{*}{\textbf{AdaGAE}} & \multirow{2}{*}{\textbf{MGAE}} & \multirow{2}{*}{\textbf{DAEGC}} & \multirow{2}{*}{\textbf{ARGA}} & \multirow{2}{*}{\textbf{DMoN}} & \multirow{2}{*}{\textbf{SDCN}} & \multirow{2}{*}{\textbf{GDCL}} & \multirow{2}{*}{\textbf{DFCN}} & \multirow{2}{*}{\textbf{Ours}} \\
                                          &                                  &                                &                                  &                                &                                 &                                &                                &                                &                                &                                &                                \\ \midrule
\multirow{4}{*}{\textbf{Cora}}            & \textbf{ACC}                     & 51.61                          & 50.06                            & 43.38                          & 70.43                           & 71.04                          & 51.70                          & 35.60                          & 70.83                          & 36.33                          & \textbf{78.10}                 \\
                                          & \textbf{NMI}                     & 26.31                          & 32.19                            & 28.78                          & 52.89                           & 51.06                          & 47.30                          & 14.28                          & 56.60                          & 19.36                          & \textbf{62.28}                 \\
                                          & \textbf{ARI}                     & 22.07                          & 28.25                            & 16.43                          & 49.63                           & 47.71                          & 30.10                          & 7.78                           & 48.05                          & 4.67                           & \textbf{61.61}                 \\
                                          & \textbf{F1}                      & 47.17                          & 53.53                            & 33.48                          & 68.27                           & 69.27                          & 57.40                          & 24.37                          & 52.88                          & 26.16                          & \textbf{72.66}                 \\ \midrule
\multirow{4}{*}{\textbf{CiteSeer}}        & \textbf{ACC}                     & 60.49                          & 54.01                            & 61.35                          & 64.54                           & 61.07                          & 38.50                          & 65.96                          & 66.39                          & 69.50                          & \textbf{70.36}                 \\
                                          & \textbf{NMI}                     & 27.17                          & 27.79                            & 34.63                          & 36.41                           & 34.40                          & 30.30                          & 38.71                          & 39.52                          & 43.90                          & \textbf{45.87}                 \\
                                          & \textbf{ARI}                     & 25.70                          & 24.19                            & 33.55                          & 37.78                           & 34.32                          & 20.00                          & 40.17                          & 41.07                          & 45.50                          & \textbf{46.96}                 \\
                                          & \textbf{F1}                      & 61.62                          & 51.11                            & 57.36                          & 62.20                           & 58.23                          & 43.70                          & 63.62                          & 61.12                          & 64.30                          & \textbf{65.96}                 \\ \midrule
\multirow{4}{*}{\textbf{Amazon-Photo}}    & \textbf{ACC}                     & 47.62                          & 67.70                            & 71.57                          & 75.96                           & 69.28                          & 24.77                          & 53.44                          & 43.75                          & 76.82                          & \textbf{81.71}                 \\
                                          & \textbf{NMI}                     & 37.83                          & 55.96                            & 62.13                          & 65.25                           & 58.36                          & 7.69                           & 44.85                          & 37.32                          & 66.23                          & \textbf{74.36}                 \\
                                          & \textbf{ARI}                     & 19.24                          & 46.20                            & 48.82                          & 58.12                           & 44.18                          & 3.81                           & 31.21                          & 21.57                          & 58.28                          & \textbf{68.40}                 \\
                                          & \textbf{F1}                      & 47.20                          & 62.95                            & 68.08                          & 69.87                           & 64.30                          & 17.98                          & 50.66                          & 38.37                          & 71.25                          & \textbf{73.51}                 \\ \midrule
\multirow{4}{*}{\textbf{ogbn-arXiv}}      & \textbf{ACC}                     & 22.67                          & \multirow{4}{*}{OOM}             & \multirow{4}{*}{OOM}           & \multirow{4}{*}{OOM}            & \multirow{4}{*}{OOM}           & 25.00                          & \multirow{4}{*}{OOM}           & \multirow{4}{*}{OOM}           & \multirow{4}{*}{OOM}           & \textbf{43.68}                 \\
                                          & \textbf{NMI}                     & 27.54                          &                                  &                                &                                 &                                & 35.60                          &                                &                                &                                & \textbf{43.73}                 \\
                                          & \textbf{ARI}                     & 12.15                          &                                  &                                &                                 &                                & 12.70                          &                                &                                &                                & \textbf{35.22}                 \\
                                          & \textbf{F1}                      & 17.58                          &                                  &                                &                                 &                                & 19.00                          &                                &                                &                                & \textbf{26.92}                 \\ \midrule
\multirow{4}{*}{\textbf{ogbn-products}}   & \textbf{ACC}                     & 20.53                          & \multirow{4}{*}{OOM}             & \multirow{4}{*}{OOM}           & \multirow{4}{*}{OOM}            & \multirow{4}{*}{OOM}           & 30.40                          & \multirow{4}{*}{OOM}           & \multirow{4}{*}{OOM}           & \multirow{4}{*}{OOM}           & \textbf{41.09}                 \\
                                          & \textbf{NMI}                     & 22.15                          &                                  &                                &                                 &                                & 42.80                          &                                &                                &                                & \textbf{50.78}                 \\
                                          & \textbf{ARI}                     & 9.87                           &                                  &                                &                                 &                                & 13.90                          &                                &                                &                                & \textbf{21.08}                 \\
                                          & \textbf{F1}                      & 12.48                          &                                  &                                &                                 &                                & 21.00                          &                                &                                &                                & \textbf{25.15}                 \\ \midrule
\multirow{4}{*}{\textbf{Reddit}}          & \textbf{ACC}                     & \multirow{4}{*}{OOM}           & \multirow{4}{*}{OOM}             & \multirow{4}{*}{OOM}           & \multirow{4}{*}{OOM}            & \multirow{4}{*}{OOM}           & 52.90                          & \multirow{4}{*}{OOM}           & \multirow{4}{*}{OOM}           & \multirow{4}{*}{OOM}           & \textbf{76.03}                 \\
                                          & \textbf{NMI}                     &                                &                                  &                                &                                 &                                & 62.80                          &                                &                                &                                & \textbf{78.91}                 \\
                                          & \textbf{ARI}                     &                                &                                  &                                &                                 &                                & 50.20                          &                                &                                &                                & \textbf{71.34}                 \\
                                          & \textbf{F1}                      &                                &                                  &                                &                                 &                                & 26.00                          &                                &                                &                                & \textbf{67.95}                 \\ \midrule
\multirow{4}{*}{\textbf{ogbn-papers100M}} & \textbf{ACC}                     & \multirow{4}{*}{OOM}           & \multirow{4}{*}{OOM}             & \multirow{4}{*}{OOM}           & \multirow{4}{*}{OOM}            & \multirow{4}{*}{OOM}           & \multirow{4}{*}{OOM}           & \multirow{4}{*}{OOM}           & \multirow{4}{*}{OOM}           & \multirow{4}{*}{OOM}           & \textbf{26.67}                 \\
                                          & \textbf{NMI}                     &                                &                                  &                                &                                 &                                &                                &                                &                                &                                & \textbf{54.92}                 \\
                                          & \textbf{ARI}                     &                                &                                  &                                &                                 &                                &                                &                                &                                &                                & \textbf{18.01}                 \\
                                          & \textbf{F1}                      &                                &                                  &                                &                                 &                                &                                &                                &                                &                                & \textbf{19.48}                 \\ \bottomrule
\end{tabular}}
\caption{Clustering performance ($\%$) of our method and nine state-of-the-art baselines. The bold values are the best results. ``OOM'' indicates that the method raise the out-of-memory failure.}
\label{compare_table}
\end{table}

Due to the limited regular paper pages, the additional compare experimental results are demonstrated in Table \ref{compare_table}. We further compare our proposed \textit{Dink-Net} with the nine baselines, including IDEC \cite{IDEC}, AdaGAE\cite{AdaGAE}, MGAE \cite{MGAE}, DAEGC \cite{DAEGC}, ARGA \cite{ARGA}, DMoN \cite{dmon}, SDCN \cite{SDCN}, GDCL \cite{GDCL}, and DFCN \cite{DFCN}. These additional experimental results further verify the superiority and scalability of our proposed \textit{Dink-Net}.

\begin{figure*}[!t]
\small
\begin{minipage}{0.139\linewidth}
\centerline{\includegraphics[width=\textwidth]{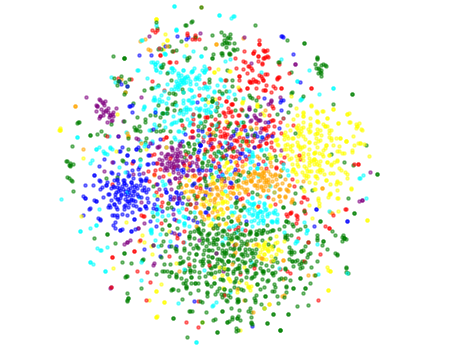}}
\vspace{3pt}
\centerline{Raw Attribute}
\end{minipage}
\begin{minipage}{0.139\linewidth}
\centerline{\includegraphics[width=\textwidth]{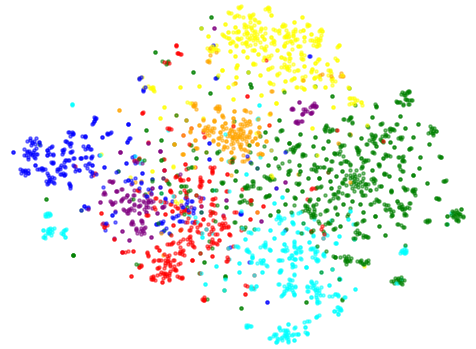}}
\vspace{3pt}
\centerline{DEC}
\end{minipage}
\begin{minipage}{0.139\linewidth}
\centerline{\includegraphics[width=\textwidth]{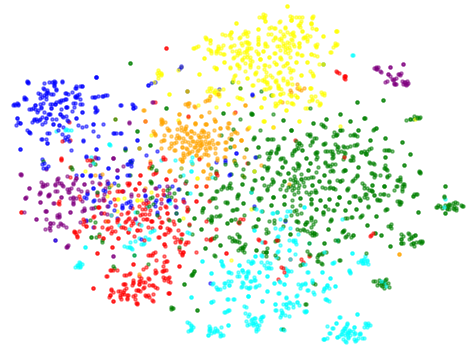}}
\vspace{3pt}
\centerline{MVGRL}
\end{minipage}
\begin{minipage}{0.139\linewidth}
\centerline{\includegraphics[width=\textwidth]{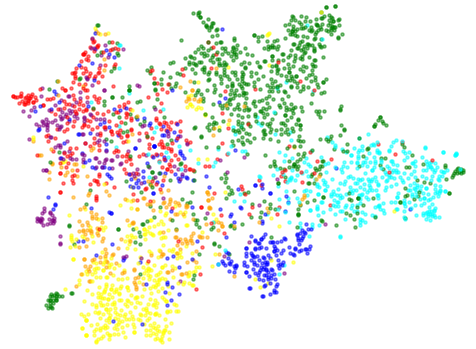}}
\vspace{3pt}
\centerline{MinCutPool}
\end{minipage}
\begin{minipage}{0.139\linewidth}
\centerline{\includegraphics[width=\textwidth]{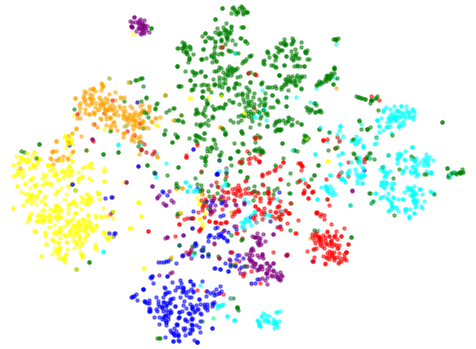}}
\vspace{3pt}
\centerline{DCRN}
\end{minipage}
\begin{minipage}{0.139\linewidth}
\centerline{\includegraphics[width=\textwidth]{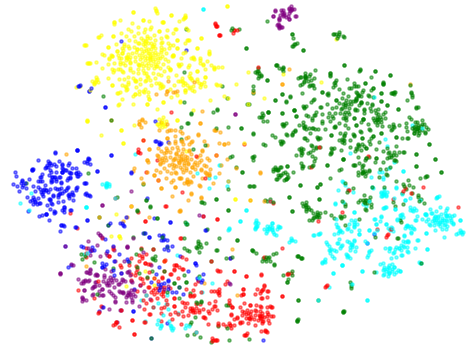}}
\vspace{3pt}
\centerline{S$^3$GC}
\end{minipage}
\begin{minipage}{0.139\linewidth}
\centerline{\includegraphics[width=\textwidth]{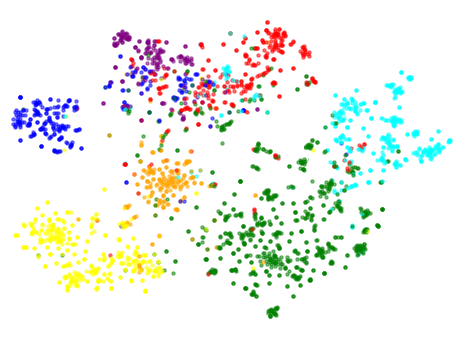}}
\vspace{3pt}
\centerline{Ours}
\end{minipage}
\caption{\textit{t}-SNE visualization of seven methods on Cora dataset.}
\label{fig:visualization}  
\end{figure*}

\subsection{Sensitivity Analyses}
\label{sec:sensitivity}
\begin{figure}[h]
\centering
\small
\begin{minipage}{0.33\linewidth}
\centerline{\includegraphics[width=1\textwidth]{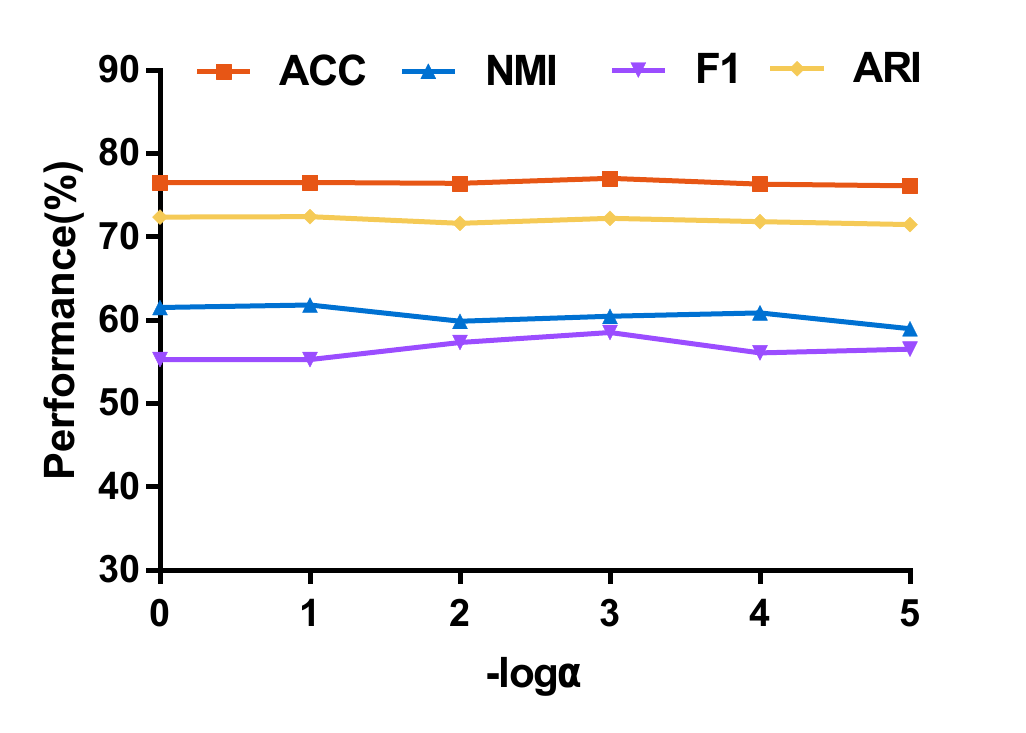}}
\centerline{(a) trade-off parameter $\alpha$ on Cora}
\vspace{3pt}
\end{minipage}
\begin{minipage}{0.33\linewidth}
\centerline{\includegraphics[width=1\textwidth]{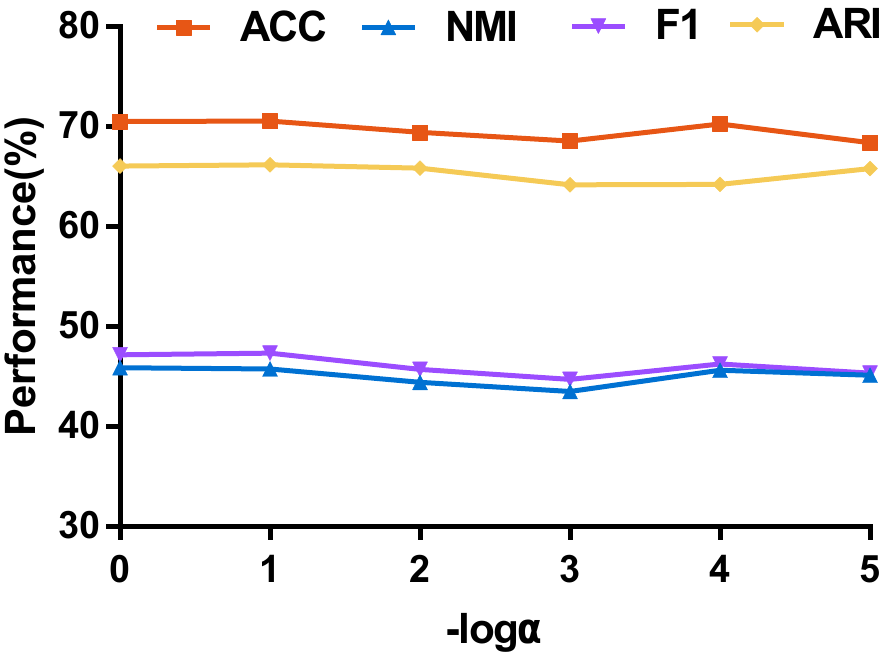}}
\centerline{(b) trade-off parameter $\alpha$ on CiteSeer}
\vspace{3pt}
\end{minipage}
\begin{minipage}{0.33\linewidth}
\centerline{\includegraphics[width=1\textwidth]{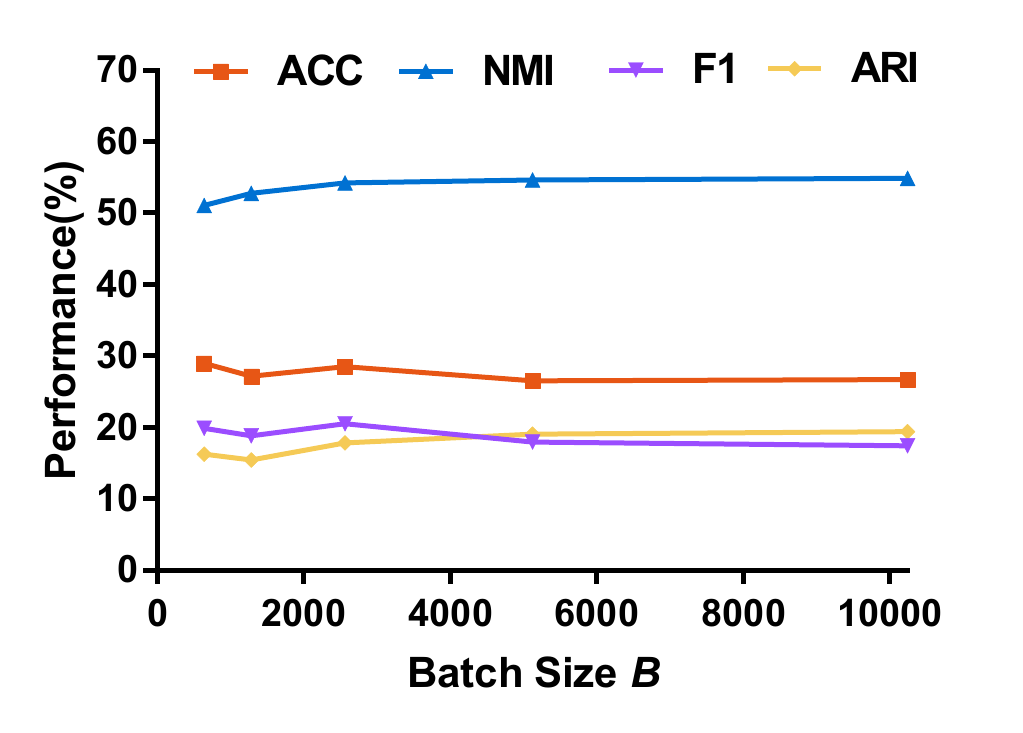}}
\centerline{(c) batch size on ogbn-papers100M}
\vspace{3pt}
\end{minipage}
\caption{Sensitivity analyses of hyper-parameters.}
\label{fig:Sensitivity}
\end{figure}

This section aims to answer \textbf{Q}5: Is the performance of the proposed method sensitive to hyper-parameters? This section analyzes the sensitivity of our proposed \textit{Dink-Net}. Firstly, we analyze the trade-off parameter $\alpha$ on Cora and CiteSeer datasets. As shown in Figure \ref{fig:Sensitivity} (a) and Figure \ref{fig:Sensitivity}(b), we find that our \textit{Dink-Net} can achieve good performance with different values of $\alpha$. Therefore it is not sensitive to $\alpha$. Secondly, we analyze another important hyper-parameter batch size $B$ on the ogbn-papers100M dataset. As shown in Figure \ref{fig:Sensitivity} (c), we find that the performance of \textit{Dink-Net} is not sensitive to batch size $B$, therefore our proposed loss functions allow our method to optimizing clustering distribution with mini-batch data even without performance drops. Besides, training model with a larger batch size will bring some extent of ACC improvement. 

\subsection{Convergence Analyses}
\begin{figure}[h]
\centering
\small
\begin{minipage}{0.45\linewidth}
\centerline{\includegraphics[width=1\textwidth]{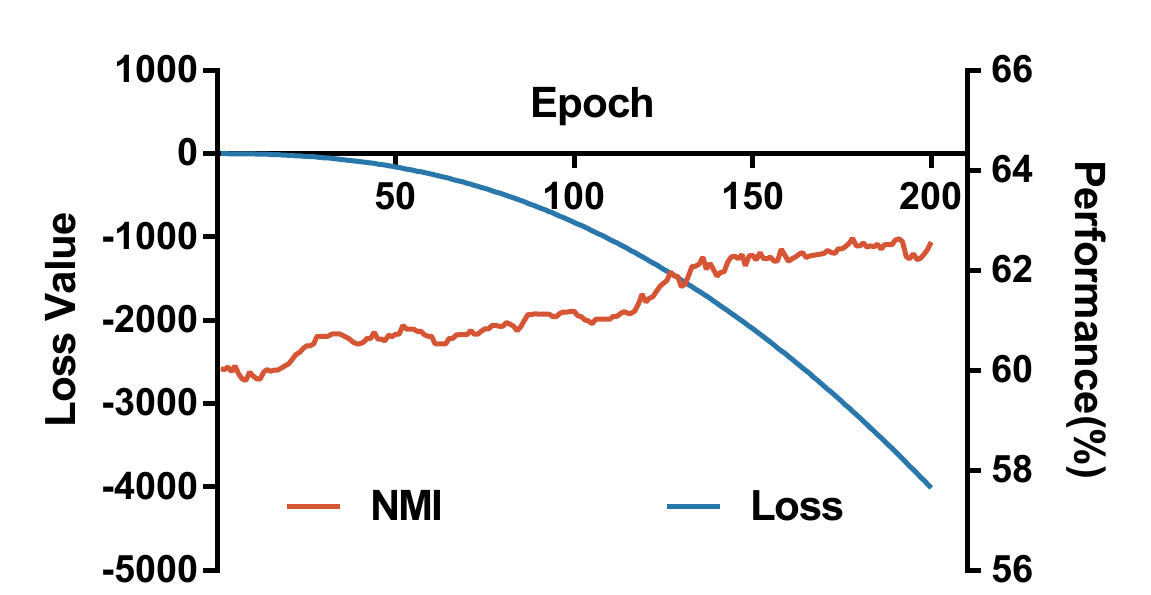}}
\centerline{(a) Cora}
\vspace{3pt}
\end{minipage}
\begin{minipage}{0.45\linewidth}
\centerline{\includegraphics[width=1\textwidth]{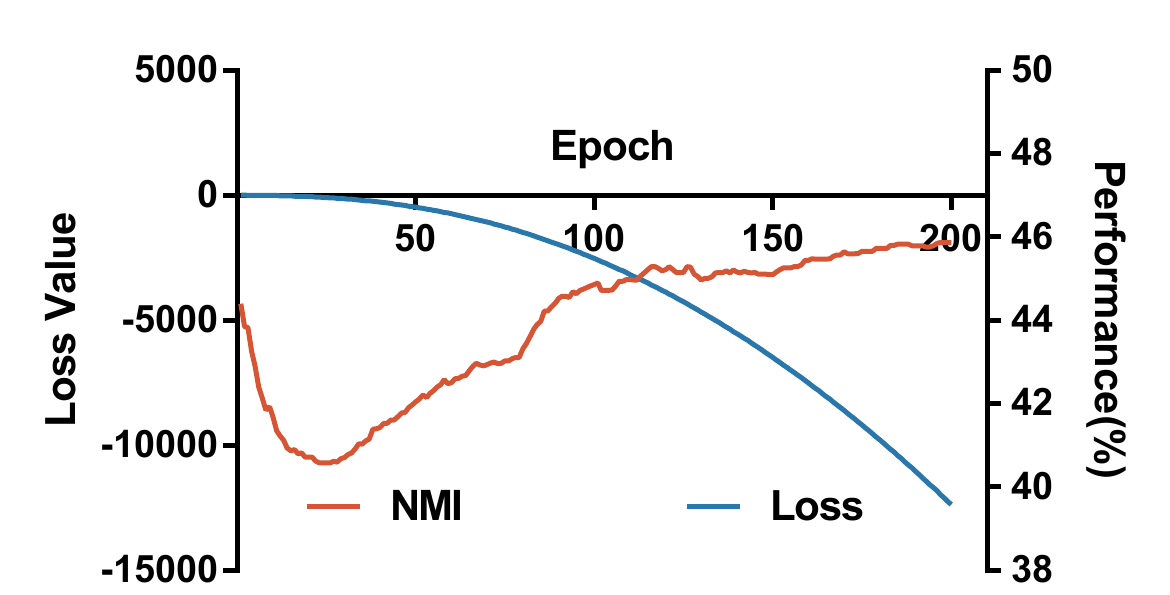}}
\centerline{(b) CiteSeer}
\vspace{3pt}
\end{minipage}
\caption{Convergence analyses on the Cora and CiteSeer datasets.}
\label{fig:convergence}
\end{figure}

This section aims to answer \textbf{Q}6: Can the proposed loss functions and the clustering performance converge well? To verify the convergence of our proposed \textit{Dink-Net}, we conduct experiments on two datasets, including Cora and CiteSeer. Specifically, as shown in Figure \ref{fig:convergence}, the NMI and loss values are recorded per epoch in the training process. From these experimental results, we observe that the loss values gradually decrease and tend to converge. Meanwhile, the clustering performance NMI increases. Therefore, our proposed method converges well.

\section{PyTorch-style Pseudo Code}
We give the PyTorch-style pseudo code of our proposed \textit{Dink-Net} in Code \ref{code}. The source codes are released on GitHub platform: https://github.com/yueliu1999/Dink-Net.

\begin{algorithm}[h]
        \centering
	\caption{PyTorch-style Pseudo Code of \textit{Dink-Net}.}
	\label{code}
	\definecolor{codeblue}{rgb}{0.25,0.5,0.5}
	\lstset{
		backgroundcolor=\color{white},
		basicstyle=\fontsize{10.2pt}{10.2pt}\ttfamily\selectfont,
		columns=fullflexible,
		breaklines=true,
		captionpos=b,
		commentstyle=\fontsize{10.2pt}{10.2pt}\color{codeblue},
		keywordstyle=\fontsize{10.2pt}{10.2pt},
	}

    \begin{lstlisting}[language=python]
    # G: attribute graph
    # F: GNN encoder
    # P: MLP projector
    # discriminate_loss: loss in Eq. (4)
    # dilation_loss: loss in Eq. (5)
    # shrink_loss: loss in Eq. (6)
    # K_Means_plus_plus: initialization of K-Means++
    # alpha: trade-off parameter
    
    # Model Pre-training Stage
    pre_optimizer = Adam(lr=pretrain_lr, parameter=Dink_net.parameters())
    for epoch in range(pretrain_epochs):
        sub_G_list = sub_graph_sampling(G)
        # batch training
        for sub_G in sub_G_list:
            sub_G_augmented = data_augmentation(sub_G)
            H = F(sub_G)
            H_ = F(sub_G_augmented)
            Z = P(H)
            Z_ = P(H_)
            dis_loss = discriminate_loss(Z.sum(-1), Z_.sum(-1))
            pre_optimizer.zero_grad()
            dis_loss.backward()
            pre_optimizer.step()
    H_all = F(G)
    C = K_Means_plus_plus(H_all)
    
    # Model Fine-tuning Stage
    fine_optimizer = Adam(lr=finetune_lr, parameter=Dink_net.parameters())
    for epoch in range(finetune_epochs):
        sub_G_list = sub_graph_sampling(G)
        # batch training
        for sub_G in sub_G_list:
            sub_G_augmented = data_augmentation(sub_G)
            H = F(sub_G)
            H_ = F(sub_G_augmented)
            Z = P(H)
            Z_ = P(H_)
            dis_loss = discriminate_loss(Z.sum(-1), Z_.sum(-1))
            dil_loss = dilation_loss(H, C)
            shr_loss = shrink_loss(H, C)
            total_loss = dil_loss + shr_loss + alpha * dis_loss
            fine_optimizer.zero_grad()
            total_loss.backward()
            fine_optimizer.step()

    # Model Inference Stage
    y_hat = []
    sub_G_list = sub_graph_sampling(G)
    for sub_G in sub_G_list:
        H = F(sub_G)
        y_hat_batch = argmin(distance(H, C))
        y_hat.append(y_hat_batch)
    return y_hat
    \end{lstlisting}
\end{algorithm}

\section{Open Resource Supports}
\subsection{Awesome Deep Graph Clustering}
This paper is supported by the Awesome Deep Graph Clustering \footnote{https://github.com/yueliu1999/Awesome-Deep-Graph-Clustering} project at GitHub. Awesome Deep Graph Clustering project summarize a comprehensive collection of the state-of-the art deep graph clustering methods, including papers, codes, and datasets. In addition, based on this GitHub project, we make a comprehensive survey about deep graph clustering \cite{liuyue_survey}. Firstly, we give the formulaic definition of deep graph clustering and introduce the milestone baselines in this field. Secondly, the taxonomy of deep graph clustering methods is presented based on four different criteria, including graph type, network architecture, learning paradigm, and clustering method. Thirdly, we carefully analyze the existing methods via extensive experiments and summarize the challenges and opportunities from five perspectives. Besides, the applications of deep graph clustering methods in four domains are presented. We hope this work can serve as a quick guide and help researchers to overcome challenges in this vibrant field.  


\subsection{A Unified Framework of Deep Graph Clustering}
In addition, this paper is supported by a unified framework of deep graph clustering\footnote{https://github.com/Marigoldwu/A-Unified-Framework-for-Deep-Attribute-Graph-Clustering} on GitHub. This GitHub project provides a practical unified framework of deep graph clustering methods. Concretely, it refactored the codes of recent state-of-the-art deep graph clustering methods to make them achieve a higher level of unification. The architecture of these codes is redesigned so that the researchers can run the open-source code efficiently. In addition, the defined tool classes and functions simplify the code and clarify the settings' configuration.


\section{URLs of Used Datasets}
This section gives the URLs of the used benchmark datasets in Table \ref{statistics_information}. 

\begin{itemize}

    \item Cora: https://docs.dgl.ai/\#CoraGraphDataset

    \item CiteSeer: https://docs.dgl.ai/\#dgl.data.CiteseerGraphDataset
    
    \item Amazon-Photo: https://docs.dgl.ai/\#dgl.data.AmazonCoBuyPhotoDataset

    \item ogbn-arxiv: https://ogb.stanford.edu/docs/nodeprop/\#ogbn-arxiv
    
    \item Reddit: https://docs.dgl.ai/\#dgl.data.RedditDataset
    
    \item ogbn-products: https://ogb.stanford.edu/docs/nodeprop/\#ogbn-products
    
    \item ogbn-papers100M: https://ogb.stanford.edu/docs/nodeprop/\#ogbn-papers100M

\end{itemize}



\bibliography{2_ref}
\bibliographystyle{icml2023}